\documentclass[10pt,journal,compsoc]{IEEEtran}
\ifCLASSOPTIONcompsoc
  \usepackage[nocompress]{cite}
\else
  \usepackage{cite}
\fi
\ifCLASSINFOpdf
\else
\fi
\usepackage{amsmath,amsfonts}
\usepackage{algorithmic}
\usepackage{algorithm}
\usepackage{array}
\usepackage[caption=false,font=normalsize,labelfont=sf,textfont=sf]{subfig}
\usepackage{textcomp}
\usepackage{stfloats}
\usepackage{url}
\usepackage{verbatim}
\usepackage{graphicx}
\usepackage{cite}
\usepackage{multirow}
\usepackage{color}

\usepackage{arydshln} 
\usepackage{amsmath,amsfonts}

\usepackage[caption=false,font=normalsize,labelfont=sf,textfont=sf]{subfig}
\usepackage{textcomp}
\usepackage{stfloats}
\usepackage{url}
\usepackage{verbatim}
\usepackage{graphicx}
\usepackage{booktabs}
\usepackage{arydshln}
\usepackage{url} 
\usepackage[colorlinks=true, linkcolor=blue, citecolor=green, urlcolor=cyan]{hyperref}
\usepackage{threeparttable}
\usepackage{ragged2e}
\usepackage{overpic}

\usepackage{hyperref}
\usepackage[table,dvipsnames]{xcolor}
\definecolor{lzh}{RGB}{153,153,255}

\begin{document}
\title{Visual Text Processing: A Comprehensive Review and Unified Evaluation}

% \title{Visual Text Meets Low-level Vision: A Comprehensive Survey on Visual Text Processing}

\author{Yan Shu, Weichao Zeng, Fangmin Zhao, Zeyu Chen, Zhenhang Li, Xiaomeng Yang\\ Yu Zhou, Paolo Rota, Xiang Bai,  Lianwen Jin, Xu-Cheng Yin,  Nicu Sebe

% \thanks{
%    • Y. Shu is with the College of Computer Science, Nankai University and with the Department of Information Engineering and Computer Science, University of Trento, Italy (e-mail: yan.shu@unitn.it).
%  }

\thanks{Y. Shu, Z. Chen, and Y. Zhou are with VCIP \& TMCC \& DISSec, College of Computer Science, Nankai University, China, and Y. Shu is also with the Department of Information Engineering and Computer Science, University of Trento, Italy (e-mail: yanshu@unitn.it; chenzeyu626@gmail.com; yzhou@nankai.edu.cn)}

\thanks{W. Zeng, F. Zhao and Z. Li are with the Institute of Information Engineering, Chinese Academy of Sciences, China, and also with the School of Cyber Security, University of Chinese Academy of Sciences, China (e-mail: zengweichao@iie.ac.cn;
zhaofangmin@iie.ac.cn; lizhenhang@iie.ac.cn)}

%\thanks{  • Z. Chen is with the College of Computer Science, Nankai University, China (e-mail: chenzeyu626@gmail.com).}

%\thanks{• Y. Zhou is with VCIP \& TMCC \& DISSec, College of Computer Science, Nankai University, China (e-mail: yzhou@nankai.edu.cn). }

\thanks{X. Yang is with the Department of Electrical and Computer Engineering, Northeastern University, USA (e-mail: yang.xiaome@northeastern.edu)}

\thanks{X. Bai is with the School of Software Engineering, Huazhong University of Science and Technology, China (e-mail: xbai@hust.edu.cn)}

\thanks{L. Jin is with the  School of Electronic and Information Engineering, South China University of Technology, China (e-mail: eelwjin@scut.edu.cn)}

\thanks{X.-C. Yin is with the Department of Computer Science and Technology and also with the Beijing Key Laboratory of Materials Science Knowledge Engineering, School of Computer and Communication Engineering, University of Science and Technology Beijing, China. (E-mail: xuchengyin@ustb.edu.cn)}

\thanks{X. Bai, L. Jin and X.-C. Yin are sorted in alphabetical order by surnames.}

\thanks{P. Rota and  N. Sebe are with the Department of Information Engineering and Computer Science, University of Trento, Italy (e-mail: paolo.rota@unitn.it; niculae.sebe@unitn.it).}

\thanks{•  Corresponding author: Y. Zhou, yzhou@nankai.edu.cn.}
}

\IEEEtitleabstractindextext{%
\begin{abstract}
\justifying
Visual text is a crucial component in both document and scene images, conveying rich semantic information and attracting significant attention in the computer vision community. Beyond traditional tasks such as text detection and recognition, visual text processing has witnessed rapid advancements driven by the emergence of foundation models, including text image reconstruction and text image manipulation. Despite significant progress, challenges remain due to the unique properties that differentiate text from general objects. Effectively capturing and leveraging these distinct textual characteristics is essential for developing robust visual text processing models. In this survey, we present a comprehensive, multi-perspective analysis of recent advancements in visual text processing, focusing on two key questions: (1) What textual features are most suitable for different visual text processing tasks? (2) How can these distinctive text features be effectively incorporated into processing frameworks? Furthermore, we introduce VTPBench, a new benchmark that encompasses a broad range of visual text processing datasets. Leveraging the advanced visual quality assessment capabilities of multimodal large language models (MLLMs), we propose VTPScore, a novel evaluation metric designed to ensure fair and reliable evaluation. Our empirical study with more than 20 specific models reveals
substantial room for improvement in the current techniques. Our aim is to establish this work as a fundamental resource that fosters future exploration and innovation in the dynamic field of visual text processing.  The relevant repository is available at \url{https://github.com/shuyansy/Visual-Text-Processing-survey}. 
\end{abstract}

}

\maketitle

\IEEEdisplaynontitleabstractindextext

% \fi
\IEEEpeerreviewmaketitle

\IEEEraisesectionheading{\section{Introduction}\label{sec:introduction}}

\IEEEPARstart{V}{isual} text, i.e., the embedded text element in images, plays an important role in various applications, including image/video retrieval \cite{fang2023uatvr}, assistive technologies for the visually impaired \cite{ezaki2004text}, scene understanding \cite{wu2024resolving}, and document artificial intelligence \cite{cui2021document}. According to the text image type, visual text can be categorized into document text and scene text. Research in this domain has primarily evolved along two branches: text spotting and text processing. Extensive research has been dedicated to text spotting, which focuses on text detection and recognition.  This field has progressed from traditional, pre-deep-learning approaches to modern deep-learning-driven paradigms. Several surveys have comprehensively reviewed these advancements~\cite{ye2014text,zhu2016scene,yin2016text,liu2019scene,lin2020review,khan2021deep,chen2021text,long2021scene}. However, while these surveys have significantly contributed to understanding text spotting, a unified review covering the entire landscape of visual text processing remains lacking.

% There has been a substantial work focusing on text spotting, originated from the era preceding deep learning to the current paradigm dominated by deep learning techniques. Numerous surveys have comprehensively reviewed these advancements \cite{ye2014text,zhu2016scene,yin2016text,liu2019scene,lin2020review,khan2021deep,chen2021text}. 
% While these surveys have significantly contributed to the field of text spotting (including detection and recognition), there remains a lack of a unified review that encompasses the entire landscape of visual text processing research.
%感觉这里的分段好奇怪

% increase in the body of work focusing on text spotting in the wild. This research evolution is mapped from the era preceding deep learning to the current paradigm dominated by deep learning techniques, which is a progression underscored by numerous studies \cite{east,crnn,db,masktext}. Thorough surveys have been documented in the literature \cite{ye2014text,zhu2016scene,yin2016text,liu2019scene,lin2020review,khan2021deep,chen2021text}, encapsulating these developments.

% Reviews by Ye et al. \cite{ye2014text} and Zhu et al. \cite{zhu2016scene} focused on image-based text detection and recognition using traditional handcrafted features. Reviews by Liu et al. \cite{liu2019scene} and Lin et al. \cite{lin2020review} shifted the emphasis toward deep learning frameworks for detecting and recognizing scene text. More recently, Chen et al. \cite{chen2021text} provided a comprehensive review of scene text recognition technologies.
 
The domain of visual text processing can be broadly categorized into text image reconstruction and text image manipulation. The former focuses on restoring and enhancing the quality of visual text, including: 1) text image super-resolution, which improves the resolution and clarity of text for low-resolution images; 2) document image dewarping, which corrects geometric distortions; 3) text image enhancement, which aims at reducing noise and enhancing image quality. In contrast, the manipulation category involves modifying visual text while preserving visual consistency, including: 1) text removal, which eliminates text from an image and restores pixels of underlying background; 2) text editing, which alters text content while preserving the original aesthetics; 3) text generation, which synthesizes text images with diverse appearances while maintaining visual authenticity. Additional related topics include text segmentation and editing detection. 

% Despite these scholarly contributions on text spotting (including detection and recognition), the literature still lacks a unified survey that integrates the full gamut of visual text processing research.

Visual text processing is essential for a wide range of practical applications. Text image enhancement and restoration primarily aim to improve the quality of low-fidelity images. This includes correcting text positioning through dewarping and enhancing readability via super-resolution or enhancement, both crucial for boosting text recognition and understanding accuracy \cite{crnn,kil2023prestu,wang2022tpsnet,wang2024partial}. Meanwhile, text image manipulation plays a vital role in privacy protection \cite{inai2014selective} through text removal, image translation \cite{fragoso2011translatar} via text editing, and augmented reality interface enhancement \cite{abu2018augmented} through text generation. With the emergence of foundation models \cite{ho2020denoising,croitoru2023diffusion,awais2025foundation}, most visual text processing models have been proposed with robust capabilities utilizing the inherent
similarities between text and general objects. 

% As illustrated in Figure \ref{fig1},
% the number of publications in this field has been steadily
% increasing over the years.

% \begin{figure}[t]
%  \setlength{\abovecaptionskip}{0cm} %调整图片标题与图距离
% \begin{center}
% \includegraphics[width=0.4\textwidth]{fig/statistic.pdf}
% \hfill
% \end{center}
% \vspace{-10pt}
% \caption{ Statistical overview of research publications in visual text processing methods. The data is collected from Google Scholar.     
% }
% \vspace{-15pt}
% \label{fig1}
% \end{figure}

% %lzh:下面这两段感觉可以在讨论一下，怪怪的
% \textcolor{lzh}{What is the main idea that these two paragraphs want to convey? It feels somewhat deviated from the central theme of text processing.}
% Visual text processing is a specialized subfield of low-level computer vision, with a particular focus on text pixels. Methodologically, it is closely related to generative artificial intelligence (AI). 
% The rapid advancement of deep learning has significantly propelled this field, particularly with the emergence of groundbreaking frameworks like Generative Adversarial Networks  \cite{goodfellow2014generative} and diffusion models \cite{ho2020denoising,croitoru2023diffusion}. These general paradigms equip visual text processing methods with robust capabilities by leveraging the inherent similarities between text and general objects, driving continuous progress over the years. As illustrated in Figure~\ref{fig1}, the number of publications in this field has been steadily increasing over the years.

Despite these advances, visual text processing presents unique challenges. Unlike general objects, visual text instances exhibit considerable variations in language, color, font, size, orientation, and shape, making its analysis inherently complex.
Therefore, analyzing these unique properties of visual text processing is crucial for optimizing future models. To address these challenges, this work provides a comprehensive, multi-perspective overview of the latest advancements in visual text processing, aiming to address two key questions: \textbf{(i) What text features are suitable for different visual text processing tasks?}
We categorize text features into four aspects: structure (layout and orientation), stroke (character glyphs), semantics (language information), and style (color and font).
\textbf{(ii) How are these distinct text features integrated into processing frameworks?} We examine a wide range of learning paradigms, including architecture designs, loss functions, learning strategies, and data representations.

Furthermore, evaluating visual text processing remains a challenge for the research community. First, \textbf{the close relationship between various visual text processing tasks has driven the development of generalist models which work in a multi-task paradigm}. For example, DocRes \cite{zhang2024docres} can handle image dewarping, de-shadowing, and deblurring tasks, while TextDiffuser \cite{chen2023textdiffuser} serves as a unified model for scene text editing and generation. Meanwhile, \textbf{evaluation results for visual text processing tasks are not always reliable \cite{lee2022surprisingly} or fair due to inconsistencies in test sets and evaluation methodologies.} Therefore, establishing a unified benchmark and proposing a standardized evaluation method that encompasses a range of visual text processing tasks would improve the fairness and reliability of evaluations. To address this, we first introduce a comprehensive benchmark VTPBench covering six different visual text processing tasks. Inspired by the strong visual quality assessment capabilities of MLLMs \cite{wu2024comprehensive,huang2024visualcritic}, we propose an effective evaluation metric, VTPScore, to ensure fair and reliable assessments. Finally, we conduct an empirical study on over 20 state-of-the-art models and identify potential areas for improvement.

% \textblue{The widespread use of text-related tasks, employing either fully annotated data under strong supervision or designing weak supervision methods, facilitates the extraction of specific text features. Furthermore, the burgeoning fields of multi-task architectures \cite{zhang2021survey} and conditional generative models \cite{mirza2014conditional,chrysos2018robust} allow for the flexible integration of various text features into different visual text processing frameworks, resulting in notable enhancements.  }    

% We begin with establishing a hierarchical taxonomy of existing works based on their processing objectives in Section~\ref{section2}. Subsequently, we conduct an in-depth discussions of seminal works, with a particular focus on the seamless integration of text features and learning paradigms in Section~\ref{section3}. Furthermore, we introduce the details of VTPBench and VTPScore  in Section~\ref{section4}.  Finally, we highlight current research challenges and suggest potential future directions in Section~\ref{section5}. The overall organization of this survey is illustrated in Figure \ref{fig2}.

\begin{figure*}[h]
 \setlength{\abovecaptionskip}{0cm} %调整图片标题与图距离
\begin{center}
\includegraphics[width=\textwidth]{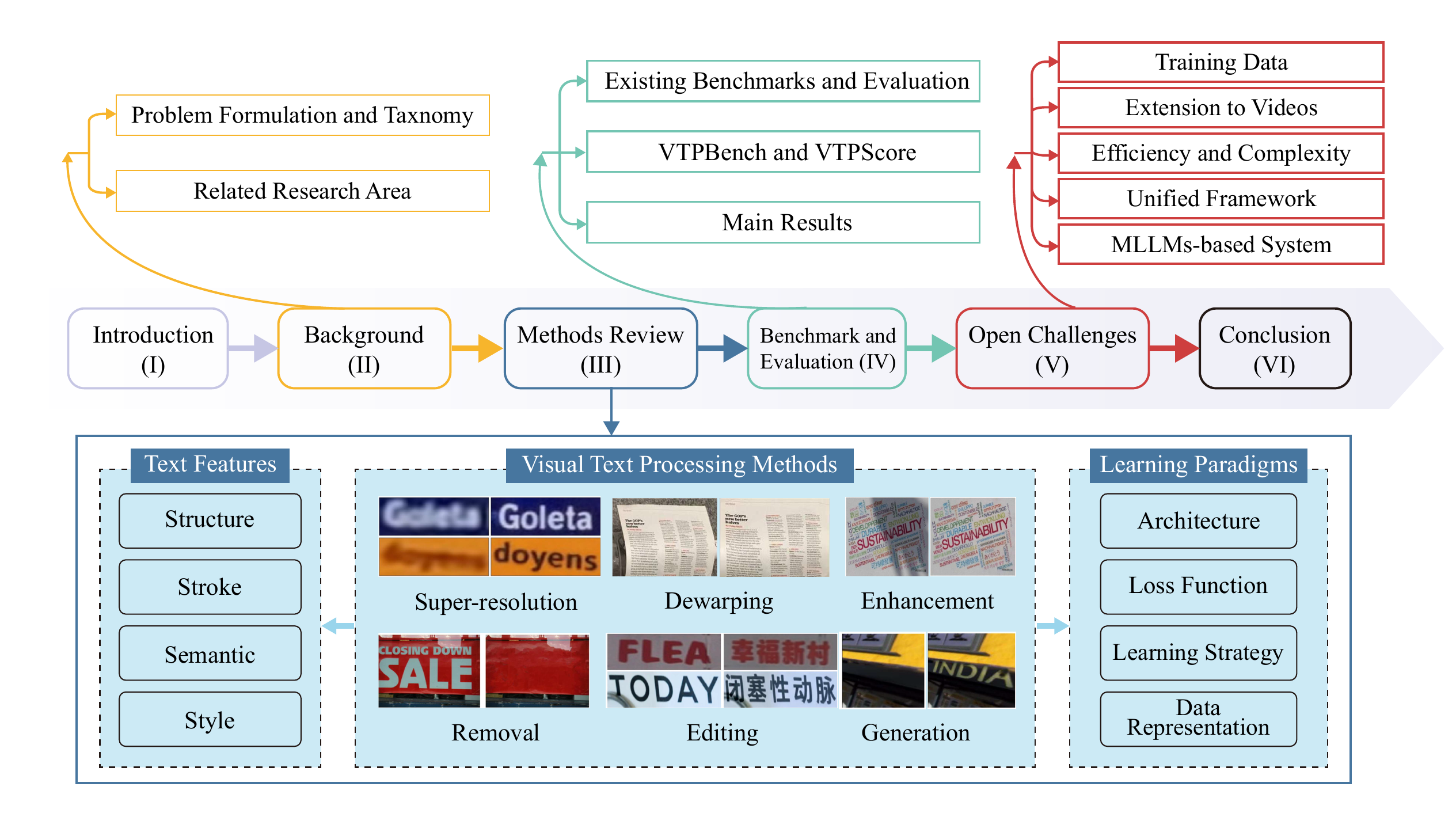}
\hfill
\end{center}
\vspace{-15pt}
\caption{Main structure of this survey.  Initially, we introduce a hierarchical taxonomy, followed by related research areas. Subsequently, we
conduct an in-depth discussion of seminar works in their distinct textual features and learning paradigms. Furthermore, we illustrate our proposed VTPBench and VTPScore for unified visual text processing evaluation. Finally, we identify open challenges for future research.} 
\vspace{-10pt}
\label{fig2}
\end{figure*}

The overall organization of this survey is illustrated in Figure \ref{fig2}. In summary, our contributions are as follows:
\begin{enumerate}
    \item Despite the existence of surveys on text detection and recognition, this is the first comprehensive review specifically focused on visual text processing to our best knowledge.
    \item We develop a multi-perspective taxonomy for visual text processing works, and highlight their distinct text features and learning paradigms. 
    \item A benchmark and an evaluation method for visual text processing are proposed, which aim to establish a unified and standardized  protocol applicable for diverse tasks.
    \item We identify and summarize open challenges in this field, providing insights on promising research directions for future exploration.
\end{enumerate}

% The organization of this survey is illustrated in Figure \ref{fig2}. Section \ref{section2} provides a concise background on problem-related taxonomy and related research areas. Section \ref{section3} thoroughly reviews representative works in this field, emphasizing their seamless integration with specific text characteristics. Section \ref{section4} examines the available datasets. Section \ref{section5} compares the reviewed works on benchmarks. Section \ref{section6} discusses the existing open challenges in the field and offers insights into potential future developments. Section \ref{section7} concludes this survey.    

\section{Background}
\label{section2}

\subsection{Problem Formulation and Taxonomy}
% Formally, let  $\boldsymbol{X}$ and $\boldsymbol{Y}$ denote the input and output spaces, respectively. Solutions in deep learning-based visual text processing typically aim to learn an optimal mapping function from pre-processed images to post-processed images, which can be mathematically represented as $f^*:\boldsymbol{X}\rightarrow\boldsymbol{Y}$. In this section, we categorize existing methods into two main areas: text image enhancement/restoration and manipulation, differentiated by the nature of  $\boldsymbol{Y}$. In subsequent section, we explore different learning paradigms according to $f^*$.

Formally, let $\boldsymbol{X}$ and $\boldsymbol{Y}$ denote the input and output spaces, respectively. Deep learning-based solutions for visual text processing typically aim to learn an optimal mapping function, mathematically represented as $f^*:\boldsymbol{X}\rightarrow\boldsymbol{Y}$. Depending on the nature of $\boldsymbol{Y}$, existing works can be broadly categorized into two main areas: text image reconstruction and text image manipulation. 

% Each area encompasses various tasks, characterized by distinct challenges and objectives. In the following subsection, we illustrate the specific contexts of $\boldsymbol{X}$ and $\boldsymbol{Y}$ \abl{or each area} and its corresponding tasks.

\subsubsection{Text Image Reconstruction}

Text images captured in natural scenes or documents often suffer from low fidelity due to factors such as low resolution, distortion, and noise interference. To address this issue, various methods have been proposed to restore the quality of text images. These methods can be further categorized into super-resolution, dewarping, and enhancement. In this context, $\boldsymbol{Y}$ should maintain the semantic consistency of $\boldsymbol{X}$, while the pixel-space distribution should be refined to align with the standards of human evaluation.

\textbf{Text Image Super-resolution.} Text image super-resolution \cite{Peyrard2015ICDAR2015CO,Nakao2019SelectiveSF,Ledig2016PhotoRealisticSI,kong2024garden,zhu2024scene,du2025instruction} aims to reconstruct high-resolution  text images $\boldsymbol{Y}$ from their low-resolution counterparts $\boldsymbol{X}$, which suffer from diverse degradations. This task significantly enhances subsequent text recognition task performance \cite{qiao2020seed,shi2016end,yu2024turning}. While sharing commonalities with general image super-resolution, it still presents unique challenges. Primarily, it is a foreground-centric task where the quality of the foreground text is paramount in evaluation, overshadowing background texture restoration. Moreover, successful restoration must not only enhance textural continuity but also preserve the semantic integrity of the text before and after restoration. This is particularly critical for languages with complex character structures like Chinese, where minor stroke discrepancies can drastically alter visual perception and lead to misinterpretation.

% Furthermore, the variability of degradative factors in real-world scenarios (such as equipment quality, lighting conditions, and compression algorithms) poses additional challenges to the generalizability of proposed methods.

% 分成两三个部分大概去讲。

\textbf{Document Image Dewarping.} Document image dewarping (DID) \cite{Dey2021LightweightDI,Das2017TheCF,zhang2009unified,meng2011metric} aims to convert distorted document images into flat images based on coordinate mapping. Uncontrollable factors such as suboptimal camera angles, improper positioning, and physical deformations of documents can severely impair the visual interpretation of document images, detrimentally affecting downstream tasks including text recognition \cite{qiao2021pimnet,wang2021pan++,liu2023spts, yang2023ipad, zhang2025linguistics}, table structure recognition \cite{shen2023divide}, and visual information extraction. In this case, $\boldsymbol{X}$ represents a distorted document image as input, while $\boldsymbol{Y}$ is the coordinate mapping between the source image and its predicted flatten version. Despite significant advancements, DID still faces substantial challenges. Current methods often rely on predefined constraints, which can lead to mode collapse in diverse application scenarios. Furthermore, while existing DID techniques generally require highly accurate ground truth for effective outcomes, current well-annotated datasets are all synthetic, leaving most real-world data unlabeled and underutilized.

\textbf{Text Image Enhancement.}
Text Image Enhancement (TIE) focuses on mitigating the negative effects, such as shadows \cite{Lin2020BEDSRNetAD,Georgiadis2023LPIOANetEH,liu2023deshadow}, stains \cite{li2024high}, blur \cite{Yang2023DocDiffDE,cicchetti2024nafdpm,zhang2024docres,wang2024docnlc,souibgui2020gan}, uneven illumination \cite{Feng2021DocTrDI,zhang2023appearance,liu2023docstormer,Wang2022UDocGANUD} and paper damage \cite{yang2024predicting,yang2024gdb,yang2023a,2019DeepOtsu,biswas2023docbinformer}, on the captured text image $\boldsymbol{X}$, aiming for a noise-free prediction $\boldsymbol{Y}$. Unlike natural image enhancement, TIE requires a meticulous approach to preserve the integrity of text structure and content. Reflecting the diversity of degradation types, research in this domain is generally divided into two primary categories: illumination removal, which addresses issues like underexposure, overexposure, and shadows; and impurity removal, a critical aspect of TIE, which concentrates on eliminating fragmented noise such as ink artifacts \cite{Huang2019DeepEraseWS}, watermarks \cite{Yang2023DocDiffDE}, stamps \cite{Yang2023MaskGuidedSE}, motion blur \cite{cicchetti2024nafdpm} and stains from damaged paper \cite{biswas2023docbinformer}. 
%Text Image Denoising (TID) is aimed at mitigating the adverse effects, such as shadows, stains, and watermarks, that impair the visual quality of text images, affecting both readability and downstream applications.% 
%Text image denoising (TID) \cite{Gangeh2021EndtoEndUD,Liu2015RestoringCD} is dedicated to reducing the negative effects, such as shadows \cite{Lin2020BEDSRNetAD,Georgiadis2023LPIOANetEH}, stains, and watermarks, on the captured text image $\boldsymbol{X}$, aiming for a noise-free prediction $\boldsymbol{Y}$. This enhancement improves readability and the performance of downstream applications like text detection \cite{east,shu2023perceiving,liao2022real}. Unlike denoising natural images, TID requires a meticulous approach to maintain the integrity of text structure and content. Reflecting the diversity of noise types, research in this domain is generally divided into two primary categories: illumination removal, addressing issues like underexposure, overexposure, and shadows; and impurity removal, a critical aspect of TID, which concentrates on eradicating fragmented noise such as ink artifacts \cite{Huang2019DeepEraseWS}, watermarks \cite{Yang2023DocDiffDE}, and stamps \cite{Yang2023MaskGuidedSE}. 
% Indeed, aside from the sequentially proposed integrated denoising model, TID is frequently addressed by a sub-module incorporated into the Document Image Dewarping (DID \textcolor{blue}{ no repetitive definition}) model. 
Whether a single, large model can effectively eliminate various types of degradation, given sufficient training data, remains an open question. 
% DocRes \cite{zhang2024docres} is an effective exploration that designs a DTSPrompt, archeving binarization, deshadowing, illumination enhancement, deblurring, and dewarping in one model.

% Traditional methods for addressing this issue have employed heuristic algorithms specifically designed to target the unique features of text images \cite{Yang2000AnAL, Gatos2006AdaptiveDD, Zhang2009AUF, Oliveira2013ShadingRO, Meng2013NonparametricIC, Bako2016RemovingSF, Kligler2018DocumentEU, Jung2018WaterFillingAE}. 

% In the deep learning era, however, more sophisticated approaches\cite{Lin2020BEDSRNetAD,Das2019DewarpNetSD, Dey2021LightweightDI, Feng2021DocTrDI, Georgiadis2023LPIOANetEH, Liu2023ShadowRO} have emerged, with the employment of text spatial context features for background estimation. Detailed discussions on these specific approaches will be presented in a later section.

% Notable independent efforts have been made in this area; for instance, Hu et al. \cite{Hu2021RecycleNetAO} and Gholamian et al. \cite{Gholamian2023HandwrittenAP} treat the separation of overlapping text as a segmentation challenge using paired data. In addressing stamp-obscured images, Yang et al. \cite{Yang2023MaskGuidedSE} introduce a specialized erasure model. Furthermore, comprehensive denoising models \cite{Gangeh2021EndtoEndUD,Yang2023DocDiffDE}, trained on datasets containing multiple types of noise, have also been developed. 

\subsubsection{Text Image Manipulation}

% Texts within images in natural scenes often require manipulation to serve various objectives, such as privacy protection, image translation, and Augmented Reality (AR)-related applications. Specific types of manipulation, including removal, editing, and generation or synthesis of text, will be detailed in this section.

%lzh: 这里写text image removal, text image editing and text image generation/synthesis 但是下面是场景是不是得统一一下，我还是建议从大的方向来  

Texts within images in natural scenes often require manipulation for various purposes, such as privacy protection, image translation, and Augmented Reality applications. Existing works mainly focus on text image removal, text image editing and text image generation. In this context, the output image $\boldsymbol{Y}$ should either maintain consistency with input image $\boldsymbol{X}$ or faithfully comply with input condition $\boldsymbol{X}$. Meanwhile, the text content should be either eliminated, modified or appended. 

%lzh:但是确实没有专门针对文档做擦除的都是顺便能做文档擦除
\textbf{Scene Text Removal.}
% \textcolor{lzh}{Text Image Removal or Scene Text Removal?}
Scene text removal (STR) \cite{nakamura2017scene,tursun2019mtrnet,tursun2020mtrnet++} is a crucial process that involves deleting text from natural images and seamlessly infilling the removed regions with contextually appropriate background pixels.  In this case, the output $\boldsymbol{Y}$ is a text-free background image. Given the widespread presence of text in images, especially on social media, STR has become essential for privacy protection. This task consists of two key sub-tasks: text localization to identify textual region and background reconstruction to replace the text pixels. Recent progress in STR methodologies has led to the development of two primary approaches: direct removal which takes only $\boldsymbol{X}$ as input and auxiliary removal, which takes both $\boldsymbol{X}$ and a binary text region segmentation mask $\boldsymbol{M}$ as input. Compared to direct removal methods, auxiliary removal methods typically achieve superior results because of their precise detection indicators.

\textbf{Scene Text Editing.} Scene text editing involves modifying text attributes, transferring styles, or altering content while ensuring seamless integration with the background, thereby preserving the overall visual coherence of the image. It can be broadly classified into two categories: style editing and content editing. In style editing, 
$\boldsymbol{Y}$ retains the same content as 
$\boldsymbol{X}$ but undergoes modifications in appearance, color, and background. Notable advancements in style editing techniques \cite{nakamura2019scene,gomez2019selective} have significantly enhanced image text processing tools, enabling smarter and more automated applications across various domains.
In contrast, content editing \cite{wu2019editing,yang2020swaptext} focuses on altering the text itself while preserving the original textual style of $\boldsymbol{X}$. In this case, $\boldsymbol{Y}$ contains modified words or characters while maintaining a consistent visual appearance. Recent works \cite{su2023scene,ji2023improving,tuo2024anytext2} attempt to solve two tasks with one model, where $\boldsymbol{Y}$ is edited on both style and content compared with $\boldsymbol{X}$.

\textbf{Scene Text Generation.} For scene text detection and recognition, scaling up training data is crucial for improving model performance. To address the challenge of labor-intensive manual annotation, text image synthesis methods \cite{zhang2019scene,balaji2022ediffi} have emerged. Beyond data augmentation, these methods also have practical applications, such as sign design and poster customization. Unlike font generalization\cite{nikolaidou2024diffusionpen,yang2024fontdiffuser,liu2024qt,yao2024geometric}, scene text generation must account for both text rendering fidelity and overall image quality.

\subsection{Related Research Areas}

This section provides an overview of scene text segmentation and editing detection, both of which are closely related to the broader field of visual text processing.

\subsubsection{Scene Text Segmentation}
Scene text segmentation \cite{ye2024hi,xie2024char} focuses on text localization using pixel-level masks, providing a more detailed detection representation compared to conventional bounding box/polygon-based text detection. This task facilitates various visual text processing methods, such as text image dewarping, scene text removal and editing, by leveraging text stroke features for fine-grained processing.

Qin et al. \cite{qin2018robust} employ the Fully Convolutional Network (FCN) to generate an initial coarse text mask, which is subsequently refined using a fully connected Conditional Random Field model. To mitigate the domain gap between synthetic and real-world text images, Bonechi et al. \cite{bonechi2020weak} develop a framework that leverages bounding box annotations of real text images to create weak pixel-level supervision. Wang et al. \cite{wang2021semi} propose a semi-supervised method that utilizes real-world data annotated with either polygon-level or pixel-level masks. Their network features a mutually reinforced dual-task architecture, consisting of a single encoder and two decoders.

Xu et al. \cite{xu2021rethinking} introduce TextSeg, a comprehensive text dataset with fine annotations, and a novel text segmentation model, TexRNet. This dataset includes 4,024 images featuring both scene and poster texts. TexRNet advances current segmentation techniques by incorporating key feature pooling and an attention module, thereby outperforming previous methods. Ren et al. \cite{ren2022looking} present a novel architecture, the Attention and Recognition enhanced Multi-scale segmentation Network, which consists of three main components: a text segmentation module, a dual perceptual decoder, and a recognition enhanced module.

% In summary, scene text segmentation, though a niche field, holds distinct importance owing to its emphasis on text strokes and characters. 

% \textcolor{blue}{This area's crucial relevance extends beyond the precision of fine-grained detection, including a range of text-focused applications.}

\subsubsection{Editing Detection} Text editing detection, or tampered text detection, plays a critical role in safeguarding privacy information.

Wang et al. \cite{wang2022detecting} propose a shared regression branch capable of capturing global semantic nuances, complemented by specialized segmentation branches to distinguish between tampered and genuine text. Additionally, their approach emphasizes frequency information extraction, as manipulations are often more detectable in the frequency spectrum than in the spatial domain. Qu et al. \cite{qu2024generalized} introduce a systematic approach to designing a more robust detection model, including a high-quality dataset curated through text editing model manipulations, a pretraining paradigm that subtly modifies the texture of selected texts within an image and a framework that considers features of both authentic and tampered text.

Compared to scene text editing detection, document text editing detection poses greater challenges due to the more subtle visual clues associated with tampering. To address this issue, Qu et al. \cite{qu2023towards} propose a novel architecture that integrates visual and frequency-domain features. Their system also incorporates a multi-view iterative decoder, specifically designed to leverage scale information for accurately detecting signs of tampering.

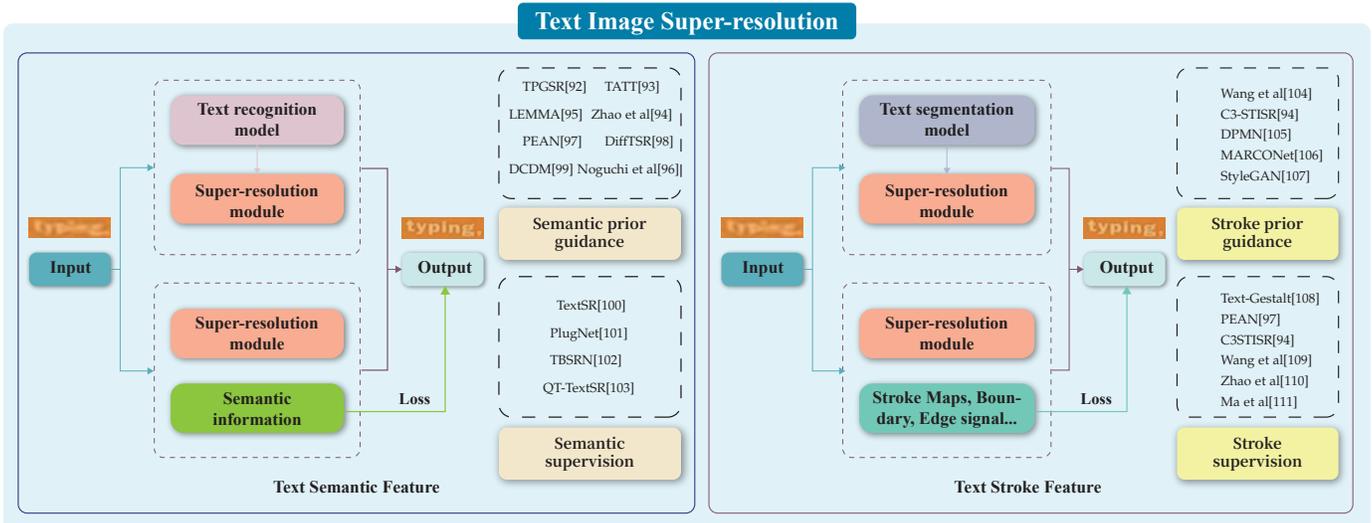
\begin{figure*}[t]
\setlength{\abovecaptionskip}{0cm}
\begin{center}
\begin{overpic}[width=\textwidth]{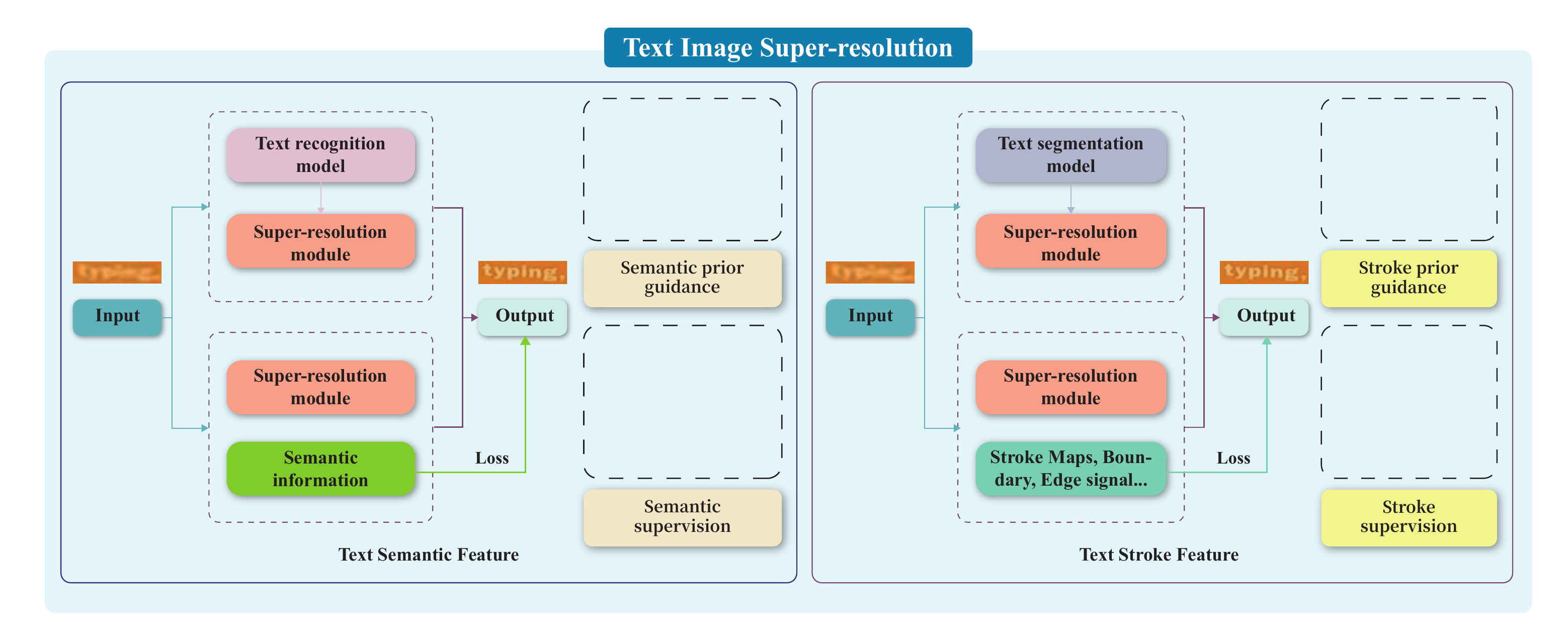}
\put(38,33){\hyperlink{cite.Ma2021TextPG}{\textcolor{Black}{\tiny TPGSR[92]}}}
\put(44,33){\hyperlink{cite.Ma2022ATA}{\textcolor{Black}{\tiny TATT[93]}}}
\put(37,31){\hyperlink{cite.guo2023towards}{\textcolor{Black}{\tiny LEMMA[95]}}}
\put(43,31){\hyperlink{cite.Zhao2022C3STISRST}{\textcolor{Black}{\tiny Zhao et al[94]}}}
\put(38,29){\hyperlink{cite.zhao2024pean}{\textcolor{Black}{\tiny PEAN[97]}}}
\put(44,29){\hyperlink{cite.zhang2024diffusion}{\textcolor{Black}{\tiny DiffTSR[98]}}}
\put(37,27){\hyperlink{cite.singh2025dcdm}{\textcolor{Black}{\tiny DCDM[99]}}}
\put(42,27){\hyperlink{cite.noguchi2024scene}{\textcolor{Black}{\tiny Noguchi et al[96]}}}
\put(40.5,17)
{\hyperlink{cite.Wang2019TextSRCT}{\textcolor{Black}{\tiny TextSR[100]}}}
\put(40,15){\hyperlink{cite.Mou2020PlugNetDA}{\textcolor{Black}{\tiny PlugNet[101]}}}
\put(40,13){\hyperlink{cite.Chen2021SceneTT}{\textcolor{Black}{\tiny TBSRN[102]}}}
\put(39.5,11){\hyperlink{cite.liu2025qt}{\textcolor{Black}{\tiny QT-TextSR[103]}}}
\put(89,32.5){\hyperlink{cite.Wang2019TextAttentionalCG}{\textcolor{Black}{\tiny Wang et al[104]}}}
\put(89,31){\hyperlink{cite.Zhao2022C3STISRST}{\textcolor{Black}{\tiny C3-STISR[94]}}}
\put(89,29.5){\hyperlink{cite.Zhu2023ImprovingST}{\textcolor{Black}{\tiny DPMN[105]}}}
\put(89,28){\hyperlink{cite.Li2023LearningGS}{\textcolor{Black}{\tiny MARCONet[106]}}}
\put(89,26.5){\hyperlink{cite.Karras2019AnalyzingAI}{\textcolor{Black}{\tiny StyleGAN[107]}}}
\put(89,17.5){\hyperlink{cite.Chen2021TextGS}{\textcolor{Black}{\tiny Text-Gestalt[108]}}}
\put(89,16){\hyperlink{cite.zhao2024pean}{\textcolor{Black}{\tiny PEAN[97]}}}
\put(89,14.5){\hyperlink{cite.Zhao2022C3STISRST}{\textcolor{Black}{\tiny C3STISR[94]}}}
\put(89,13){\hyperlink{cite.Wang2020Scene}{\textcolor{Black}{\tiny Wang et al[109]}}}
\put(89,11.5){\hyperlink{cite.Zhao2022C3STISRST}{\textcolor{Black}{\tiny Zhao et al[110]}}}
\put(89,10){\hyperlink{cite.Ma2023ABF}{\textcolor{Black}{\tiny Ma et al[111]}}}
\end{overpic}
\hfill
\end{center}
\vspace{-10pt}
\caption{Text semantic and stroke features are utilized for \textbf{text image super-resolution} with prior (pre-trained models) or supervision (auxiliary loss for model outcome).} 
\vspace{-15pt}
\label{fig:sr_pipeline}
\end{figure*}

\section{Methods Review}
\label{section3}
In this section, we review key methods across various visual text processing tasks. We focus on two main aspects: first, exploring essential text-related features such as structure, stroke, semantics, style, and layout; second, highlighting the distinct learning paradigms that underpin these approaches. 

\subsection{Text Image Super-resolution}
Traditional image super-resolution methods mainly focus on reconstructing fine textures of natural images, leading to low generalization to text images.  Existing works mostly utilize text-related features to guide networks to pay more attention to textual content, primarily relying on semantic and stroke features, as shown in Figure \ref{fig:sr_pipeline}.

\subsubsection{Text Semantic Feature}
 Existing methods typically incorporate text semantic features to enhance the performance of text super-resolution methods, employing two primary approaches: semantic prior guidance as model input and text recognition supervision for model outcome.

\textbf{Semantic Prior Guidance.} In the field of text image super-resolution, Text Prior refers to the probabilistic sequence of text obtained from text recognition models. TPGSR \cite{Ma2021TextPG} pioneers the utilization of pre-trained text recognition models as text prior generators. Following the extraction of text prior information, the framework employs a Text Prior Transformer to derive text prior features, which subsequently guide the super-resolution process.

This approach gains significant attention in the field, with numerous subsequent studies building upon and refining it. TATT \cite{Ma2022ATA} develops a transformer-based module to synchronize text priors with spatially-deformed text images, ensuring accurate feature alignment. Zhao et al. \cite{Zhao2022C3STISRST} introduce the C3-STISR, a triple clue-assisted network. This network leverages recognition, visual, and linguistic cues to enhance super-resolution. LEMMA \cite{guo2023towards} proposes a novel framework that explicitly enhances text location features through a dedicated location enhancement module, improving the robustness of scene text image super-resolution under complex spatial deformations. Noguchi et al. \cite{noguchi2024scene} represent the pioneering effort applying text-conditional DMs to scene text image super-resolution. PEAN \cite{zhao2024pean} and  DiffTSR \cite{zhang2024diffusion} adopt diffusion models to refine and enhance text priors. Additionally, DCDM \cite{singh2025dcdm} proposes a latent text diffusion model to generate text prior, eliminating the need for a text recognizer during inference. 

\textbf{Semantic Supervision.} Several approaches employ loss functions to guide the network in learning semantic features of text, ensuring that the super-resolution process emphasizes textual regions within the images. For instance, TextSR \cite{Wang2019TextSRCT} proposes a generative model-based super-resolution framework and integrates a text recognition branch for multi-task learning. By incorporating text recognition loss, they jointly optimize both recognition and super-resolution tasks, enhancing the overall performance. PlugNet \cite{Mou2020PlugNetDA} introduces a pluggable SR unit within a multi-task framework to simultaneously perform super-resolution and text recognition. TBSRN \cite{Chen2021SceneTT} utilizes a pre-trained Transformer to construct a Content-Aware Module, which predicts text sequences and computes a content-aware loss using a variational autoencoder (VAE) to enhance the discriminative quality of the reconstructed text regions in super-resolution images. QT-TextSR \cite{liu2025qt} also leverages text recognition supervision by using a Query-aware Transformer to guide the super-resolution process, ensuring that the enhanced images emphasize textual regions.

\subsubsection{Text Stroke Feature}
The quality of strokes plays an essential role in text recognition. This significance extends to the field of text image super-resolution, where accurately recovering and enhancing stroke-level details is crucial for improving the recognizability of text in low-resolution images.

\textbf{Stroke Prior Guidance.} Glyph maps, segmentation maps, and other stroke-based representations provide rich stroke-level priors, which are widely used to enhance the network's focus on fine-grained stroke details of text characters. Wang et al. \cite{Wang2019TextAttentionalCG} incorporate text/non-text segmentation maps as input to provide stroke prior, and design a text spatial attention mechanism to guide the model to focus more on text regions rather than image backgrounds. C3-STISR \cite{Zhao2022C3STISRST} and DPMN \cite{Zhu2023ImprovingST} renders the text recognition results of low-resolution images into glyph maps to represent the structural details of text content. MARCONet \cite{Li2023LearningGS} utilizes StyleGAN \cite{Karras2019AnalyzingAI} to capture a wide range of structural text variations, leveraging generative structure priors for accurate text image restoration.

\textbf{Stroke  Supervision.} Text-Gestalt \cite{Chen2021TextGS} proposes a strategy that deconstructs characters into strokes and uses stroke-level attention maps from an auxiliary recognizer to guide the super-resolution. It also introduces a Stroke-Focus Module Loss to align the stroke-level attention maps of the super-resolved images with the ground truth, ensuring finer recovery of stroke details. This loss is further adopted in subsequent works, such as PEAN \cite{zhao2024pean} and C3-STISR \cite{Zhao2022C3STISRST}, demonstrating its effectiveness in enhancing stroke-level text image super-resolution. Some works use boundary-aware losses to sharpen edges, providing the network with supervision on text stroke. Wang et al. \cite{Wang2020Scene} introduce the first real-world text super-resolution paired dataset, TextZoom, and propose the Gradient Profile Loss, which leverages gradient fields to supervise the recovery of character edges, thereby generating sharper text images.  Zhao et al. \cite{Zhao2021SceneTI} utilize the Sobel operator to compute edge loss. Additionally, Ma et al. \cite{Ma2023ABF} propose a real-world Chinese-English benchmark dataset and develop an edge-aware learning method supervised by text edge maps.

% \begin{figure}[t]
%  \setlength{\abovecaptionskip}{0cm} %调整图片标题与图距离
% \begin{center}
% \includegraphics[width=0.5\textwidth]{fig/did.pdf}
% \hfill
% \end{center}
% \vspace{-10pt}
% \caption{Text structure features are utilized for 
% \textbf{document image dewarping}  with two-stage or end-to-end learning.} 
% \vspace{-15pt}
% \label{fig:did_pipeline}
% \end{figure}

\begin{figure}[t]
\setlength{\abovecaptionskip}{0cm}
\begin{center}
\begin{overpic}[width=0.5\textwidth]{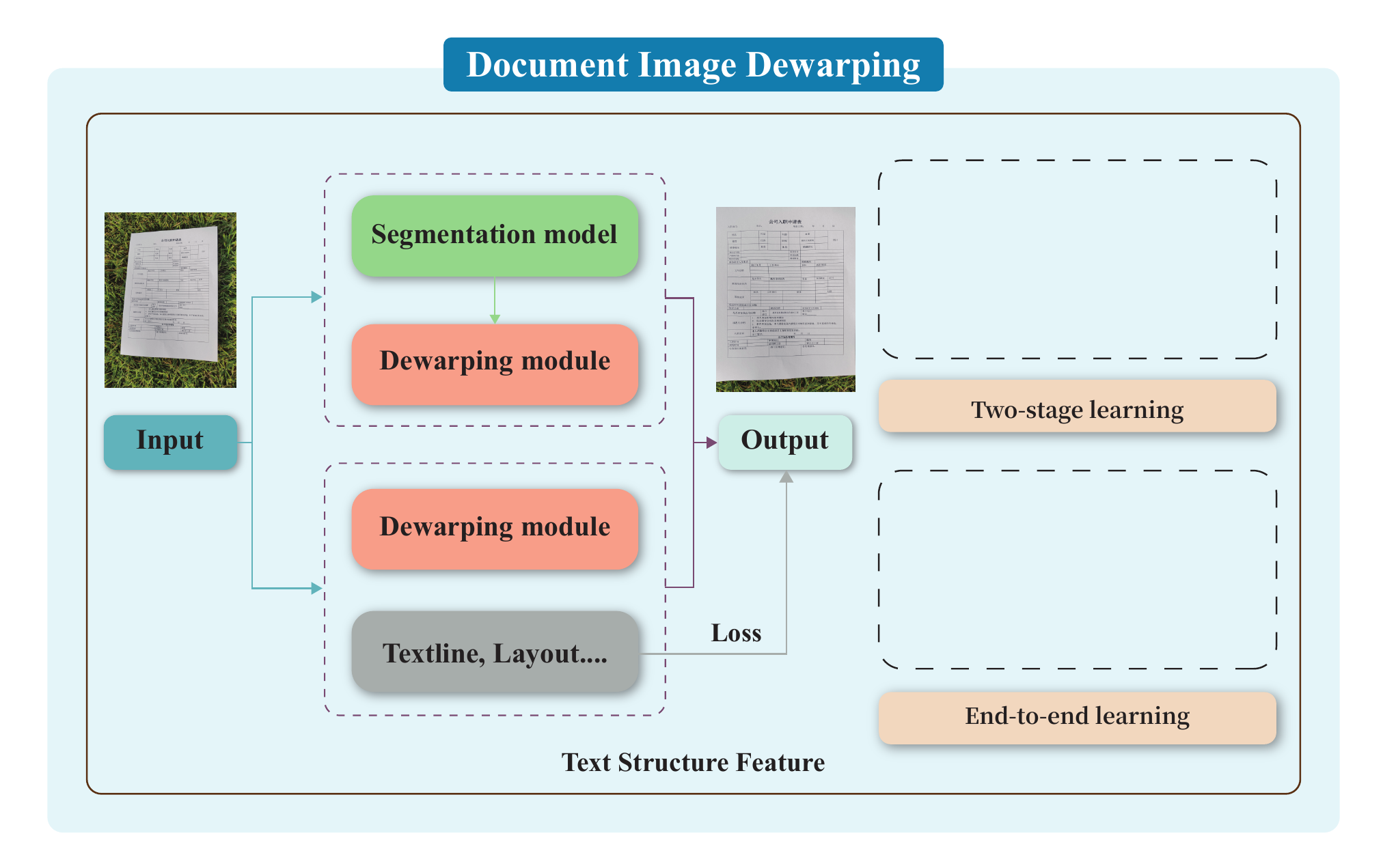}

    \put(68,48){\hyperlink{cite.Feng2021DocTrDI}{\textcolor{Black}{\tiny DocTr[55]}}}
    
    \put(80,48){\hyperlink{cite.Xie2020DewarpingDI}{\textcolor{Black}{\tiny Xie et al[113]}}}
    \put(68,45){\hyperlink{cite.Zhang2022MariorMR}{\textcolor{Black}{\tiny Marior[112]}}}
        \put(80,45){\hyperlink{cite.zhang2024docres}{\textcolor{Black}{\tiny DocRes[23]}}}
        \put(67,42){\hyperlink{cite.Feng2022GeometricRL}{\textcolor{Black}{\tiny DocGeoNet[114]}}}
  \put(82,42){\hyperlink{cite.Feng2023DeepUD}{\textcolor{Black}{\tiny DocTr++[115]}}}
  \put(68,39){\hyperlink{cite.liu2023rethinking}{\textcolor{Black}{\tiny DRNet[116]}}}
    \put(80,39){\hyperlink{cite.tang2024efficient}{\textcolor{Black}{\tiny Tang et al[117]}}}
 \put(66,24){\hyperlink{cite.Das2019DewarpNetSD}{\textcolor{Black}{\tiny DewarpNet[120]}}}
  \put(82,24){\hyperlink{cite.verhoeven2023uvdoc}{\textcolor{Black}{\tiny UVDoc[122]}}}
  \put(67,20){\hyperlink{cite.xu2022document}{\textcolor{Black}{\tiny Xu et al[121]}}}
    \put(80,20){\hyperlink{cite.Feng2022GeometricRL}{\textcolor{Black}{\tiny DocGeoNet[114]}}}
      \put(72,16){\hyperlink{cite.Li2023LayoutAwareSD}{\textcolor{Black}{\tiny LA-DocFlatten[123]}}}
  
\end{overpic}
\hfill
\end{center}
\vspace{-10pt}
\caption{Text structure features are utilized for 
\textbf{document image dewarping}  with two-stage or end-to-end learning.}
\vspace{-15pt}
\label{fig:did_pipeline}
\end{figure}
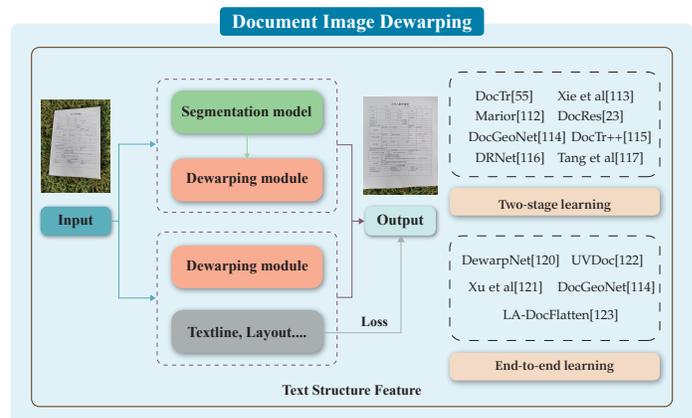

\subsection{Document Image Dewarping}

Document images serve as one of the primary carriers of text, containing rich textual information. Document Image Dewarping aims to eliminate geometric distortions to enhance readability and improve the OCR accuracy of the text within the document.
In most related works, text structure features play a crucial role in guiding the dewarping process (Figure \ref{fig:did_pipeline}).

\subsubsection{Text Structure Feature}
Text Structure Features in DID typically include text lines, document layout, boundaries, and 3D information. For example, a flat document should have complete boundaries, horizontal text lines, and a rectangular layout.
The extraction and utilization of text structure features typically follow two-stage learning or end-to-end learning paradigm.

\textbf{Two-stage Learning.}  Due to the challenges of annotating dewarping datasets and the lack of high-quality training data, many methods adopt a two-stage learning approach, where an explicit text structure feature extractor is first trained using more external data. In the second stage, they fuse these extracted features to facilitate the dewarping process. Several works, including DocTr \cite{Feng2021DocTrDI} and Marior \cite{Zhang2022MariorMR}, along with other studies \cite{Xie2020DewarpingDI,Feng2022GeometricRL,Feng2023DeepUD,liu2023rethinking,tang2024efficient,zhang2024docres}, first train a segmentation model to extract boundary information from documents. The extracted segmentation mask is then multiplied with the original image to remove background information, thereby decoupling document localization from document dewarping. The processed image is subsequently fed into a regression network to predict the final flow mapping for dewarping. RDGD \cite{Jiang2022RevisitingDI} first trains a document boundary extraction model and a text line segmentation model, and solves an optimization problem with their proposed grid regularization.
Li et al. \cite{li2023foreground} utilize a well-trained boundary segmentation model and a text line detection model to extract their intermediate structural features, followed by cross-attention operations to fuse these features with document features, assisting in dewarping and demonstrating significant improvement in OCR accuracy with more horizontal text lines. 

\textbf{End-to-end Learning.} End-to-end learning can reduce error accumulation and achieve a more streamlined training process. DewarpNet \cite{Das2019DewarpNetSD} and Xu et al. \cite{xu2022document} design two cascaded UNet-like regression networks, where the first sub-network receives the original document and learns to predict 3D features, while the second network receives the 3D features and learns the mapping from 3D shape to 2D flow map. These networks divide the dewarping task into manageable parts, simplifying each step. UVDoc \cite{verhoeven2023uvdoc} incorporates two output heads after the feature extractor, enabling simultaneous prediction of both 3D shape and 2D flow mapping. The prediction of 3D shape serves as an auxiliary task during training. DocGeoNet \cite{Feng2022GeometricRL} adopts a hybrid architecture, using a transformer-based sub-network for 3D shape prediction and a CNN-based sub-network for text line learning. Their representations are concatenated and passed through a decoder, which predicts the final rectification. These two sub-networks and decoder are end-to-end optimized. LA-DocFlatten \cite{Li2023LayoutAwareSD} designs a dual-decoder network to perform document layout analysis and global document dewarping simultaneously. Subsequently, it dewarps each layout within the document, achieving finer-grained rectification.

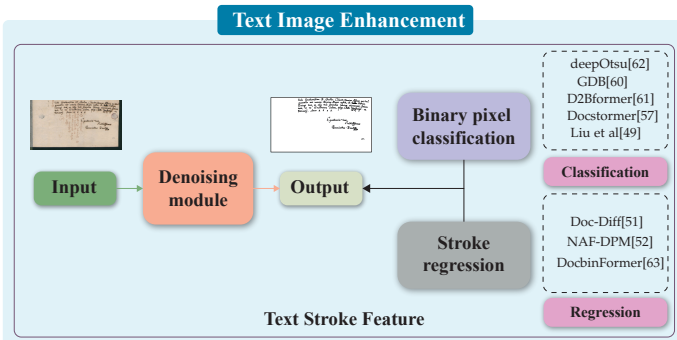
\begin{figure}[t]
\setlength{\abovecaptionskip}{0cm}
\begin{center}
\begin{overpic}[width=0.5\textwidth]{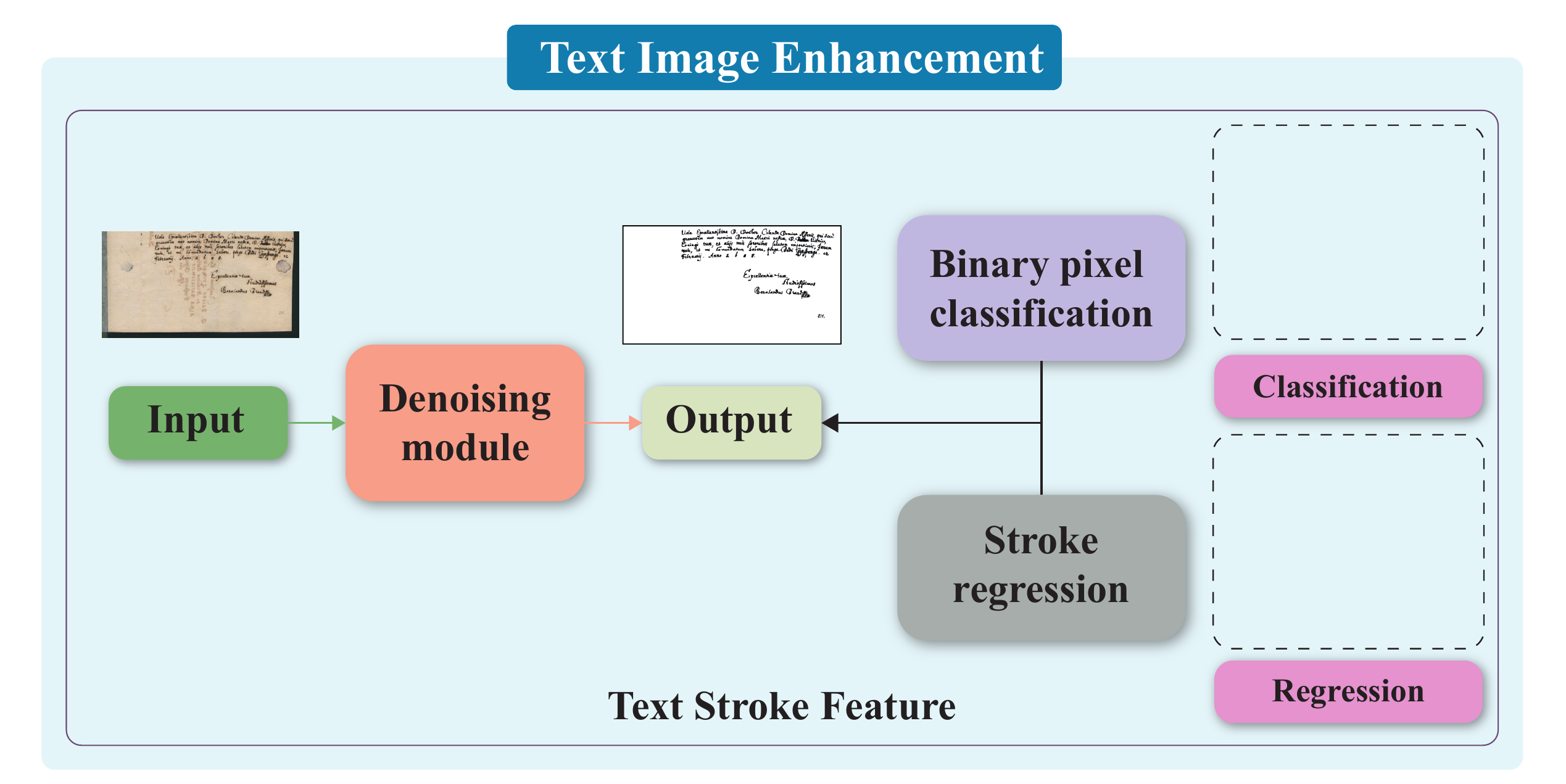}
    % 小字体+漂亮颜色
    \put(83,41){\hyperlink{cite.2019DeepOtsu}{\textcolor{Black}{\tiny deepOtsu[62]}}}
    \put(84,38.5){\hyperlink{cite.yang2024gdb}{\textcolor{Black}{\tiny GDB[60]}}}
\put(82.5,36){\hyperlink{cite.yang2023a}{\textcolor{Black}{\tiny D2Bformer[61]}}}
\put(82.5,33.5){\hyperlink{cite.liu2023docstormer}{\textcolor{Black}{\tiny Docstormer[57]}}}
\put(83,31){\hyperlink{cite.liu2023deshadow}{\textcolor{Black}{\tiny Liu et al[49]}}}
\put(83,18){\hyperlink{cite.Yang2023DocDiffDE}{\textcolor{Black}{\tiny Doc-Diff[51]}}}
\put(82.5,15){\hyperlink{cite.cicchetti2024nafdpm}{\textcolor{Black}{\tiny NAF-DPM[52]}}}
\put(81,12){\hyperlink{cite.biswas2023docbinformer}{\textcolor{Black}{\tiny DocbinFormer[63]}}}
\end{overpic}
\hfill
\end{center}
\vspace{-10pt}
\caption{\textbf{Text image enhancement} methods leverage text stroke feature to perform classification or regression training paradigm.} 
\vspace{-15pt}
\label{fig:tid_pipeline}
\end{figure}

\subsection{Text Image Enhancement}

Text image enhancement focuses on removing background noise, such as shadow and ink bleed-through, while preserving text integrity. Therefore, both text perception and degradation modeling play a crucial role in this task. Text stroke features and text semantic features are frequently used, as demonstrated in Figure \ref{fig:tid_pipeline}.

\subsubsection{Text Stroke Feature}

To extract text stroke features from text images, some methods model it as a classification problem, which classifies each pixel as either part of the text region or the background. Other approaches model it as a regression problem, treating the stroke binary mask as an image and predict it directly.

\textbf{Classification.} DeepOtsu \cite{2019DeepOtsu} first utilizes an iterative UNet-like network to predict degradation values of the image, which are then subtracted from the original image to obtain a clean version. The clean image is subsequently classified using an SVM to produce a binary result. GDB \cite{yang2024gdb} enhances binary image prediction by combining multiple inputs, including the degraded image, a coarse binary image from Otsu’s algorithm, and Sobel gradients. The model predicts both a refined binary image and an edge image, employing classification loss and L1 regression loss to stabilize training, reduce ambiguity, and enforce pixel-level consistency. $D^2BFormer$ \cite{yang2023a} combines Transformer and CNN architectures to extract features that are sensitive to both global and local contexts, aiding in segmentation. In addition to the binary classification of each pixel, dice loss is introduced to mitigate the issue of imbalance between the number of foreground and background pixels. Docstormer \cite{liu2023docstormer} designs a degradation-perception network to predict degradation information, such as ink bleed-through stroke masks in images. Multi-scale degradation features are then fed into the decoder to assist in noise removal. Moreover, Liu et al. \cite{liu2023deshadow} design an adaptive dynamic strategy to predict the threshold for each pixel, thereby obtaining the classification results of auxiliary pixels. By combining the predicted mask with the background, they achieve effective shadow removal.

\textbf{Regression}-based methods treat the mask of text strokes as an image to be predicted, making them adaptable to a wider range of structures and tasks. DocDiff \cite{Yang2023DocDiffDE} and NAF-DPM \cite{cicchetti2024nafdpm} utilize the same model architecture but are trained with different parameters to perform text image deblurring and binarization, respectively. DocBinFormer \cite{biswas2023docbinformer} employs a two-level transformer encoder to effectively capture both global and local feature representations from the input images, which improves binarization performance for both system-generated and handwritten document images. All these methods utilize MSE Loss for supervision, ensuring accurate stroke reconstruction and noise reduction.

\subsubsection{Text Semantic Feature}

Since text image enhancement typically targets dense text scenarios like document images, the large volume of text presents a challenge for effectively extracting text semantic features. Currently, NAF-DPM \cite{cicchetti2024nafdpm} is the only work that utilizes text semantic features to achieve enhancement. NAF-DPM uses a diffusion model to predict the residual between the enhanced result from the preliminary coarse regression network and the target image. To ensure the preservation of text semantic features, it calculates the CTC loss between the target image and the predicted image, guiding the network to refine textual details. The final restored image achieves a lower CER compared to previous methods, demonstrating significant improvements in OCR results.

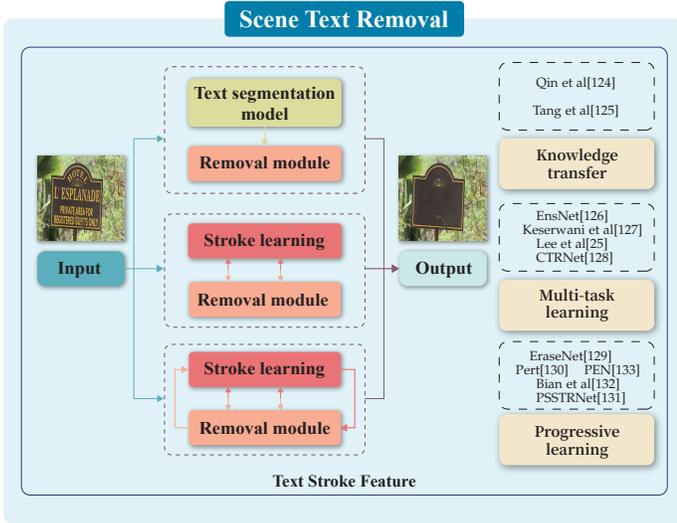
\begin{figure}[t]
\setlength{\abovecaptionskip}{0cm}
\begin{center}
\begin{overpic}[width=0.5\textwidth]{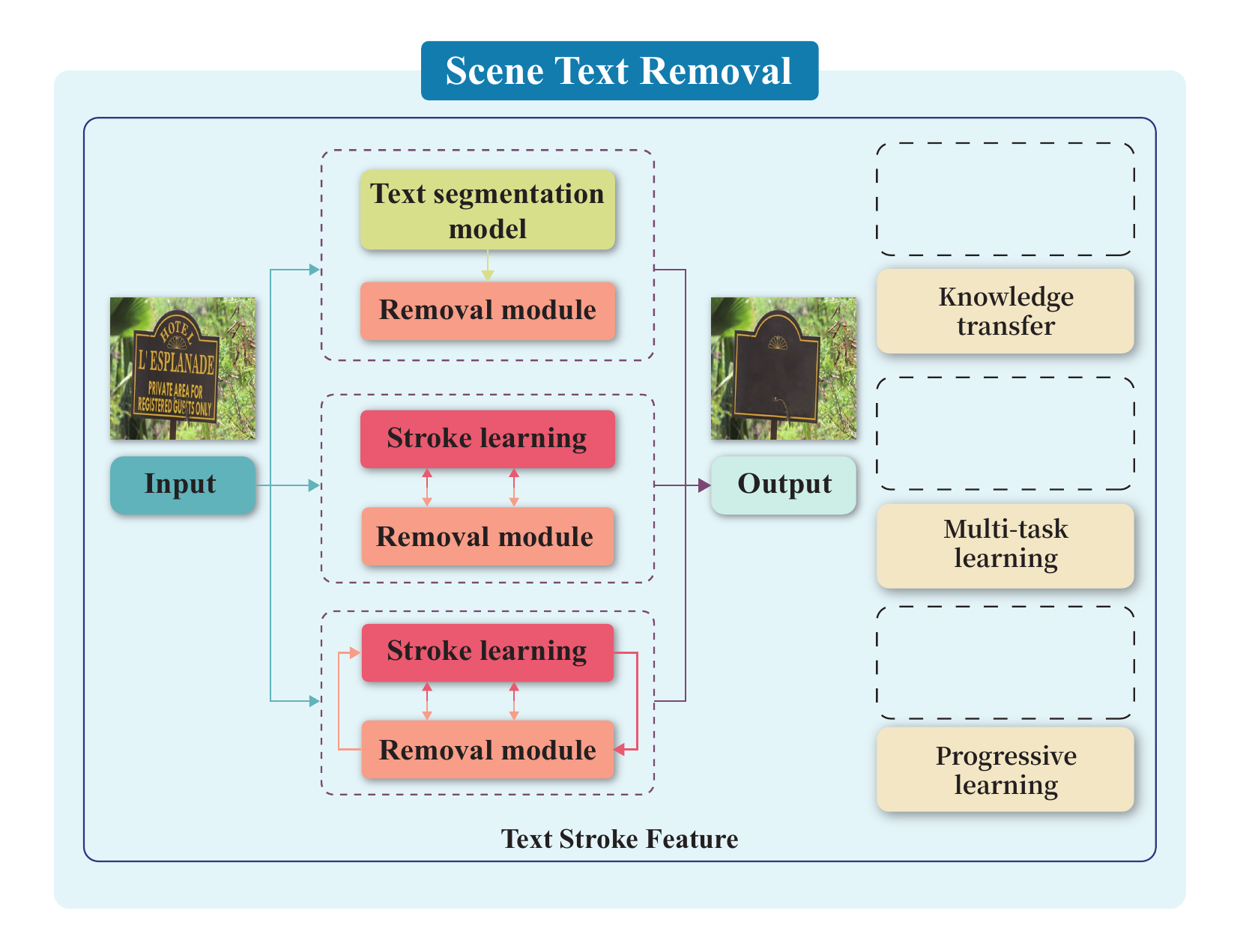}
    % 小字体+漂亮颜色
    \put(78,64){\hyperlink{cite.qin2018automatic}{\textcolor{Black}{\tiny Qin et al[124]}}}
    \put(77.5,60){\hyperlink{cite.tang2021stroke}{\textcolor{Black}{\tiny Tang et al[125]}}}
        \put(78,44.5){\hyperlink{cite.zhang2019ensnet}{\textcolor{Black}{\tiny EnsNet[126]}}}
      \put(76,42.5){\hyperlink{cite.keserwani2021text}{\textcolor{Black}{\tiny Keserwani et al[127]}}}
         \put(78,40.5){\hyperlink{cite.lee2022surprisingly}{\textcolor{Black}{\tiny Lee et al[25]}}}
             \put(78,38.5){\hyperlink{cite.liu2022don}{\textcolor{Black}{\tiny CTRNet[128]}}}
            \put(77,24){\hyperlink{cite.liu2020erasenet}{\textcolor{Black}{\tiny EraseNet[129]}}}
              \put(75,22){\hyperlink{cite.wang2021pert}{\textcolor{Black}{\tiny Pert[130]}}}
               \put(85,22){\hyperlink{cite.du2023progressive}{\textcolor{Black}{\tiny PEN[133]}}}
                           \put(78,18){\hyperlink{cite.lyu2022psstrnet}{\textcolor{Black}{\tiny PSSTRNet[131]}}}
        \put(78,20){\hyperlink{cite.bian2022scene}{\textcolor{Black}{\tiny Bian et al[132]}}}
\end{overpic}
\hfill
\end{center}
\vspace{-10pt}
\caption{Text stroke features are utilized for \textbf{scene text removal} task with knowledge transfer, multi-task learning and progressive learning.} 
\vspace{-15pt}
\label{fig:removal_pipeline}
\end{figure}

\subsection{Scene Text Removal}
In order to erase character glyphs while preserving the background, text stroke features are fully utilized as fine-grained guidance in scene text removal.

\subsubsection{Text Stroke Feature}
Existing methods typically apply knowledge transfer, multi-task learning, or progressive learning strategies to accurately represent text stroke information, enhancing the performance of text removal, as shown in Figure~\ref{fig:removal_pipeline}.

\textbf{Knowledge Transfer.} Text stroke predictions can be obtained as either primary outputs or by-products of text segmentation and detection models. Therefore, Qin et al. \cite{qin2018automatic} and Tang et al. \cite{tang2021stroke} apply pretrained text detection models to segment text regions before removal, facilitating more straightforward text stroke extraction. The generated stroke masks assist in reconstructing the background.  \cite{tang2021stroke} implements a sequential process, while \cite{qin2018automatic} uses a parallel decoding strategy to integrate stroke characteristics effectively during inpainting.

\textbf{Multi-task Learning.} Many works have been done to design a synergetic framework for efficiently learning stroke representation and predicting removing results. EnsNet \cite{zhang2019ensnet} builds GAN to simultaneously learn the multi-scale background information and local text region features. Keserwani et al. \cite{keserwani2021text} address this challenge by introducing a symmetric line character representation to improve stroke feature prediction. A specialized mask loss is employed to direct the network in learning essential features. Lee et al. \cite{lee2022surprisingly} further extract text stroke region and text stroke surrounding region with weakly supervised learning. They use a gated attention mechanism to adjust confidence levels across these regions, leading to more precise segmentation of text strokes. CTRNet \cite{liu2022don} devises a low-level contextual guidance block to capture image structural details, alongside a high-level contextual guidance block focusing on semantic aspects of the latent feature space. Moreover, they incorporate a feature content modeling block to blend the immediate pixels around text areas with the broader background, thereby minimizing texture inconsistencies in complex settings.

\textbf{Progressive Learning.} An intuitive way to enhance text stroke representation is a progressive strategy. EraseNet \cite{liu2020erasenet} establishes a coarse-to-fine pipeline, progressively erasing text regions before refining stroke-level details. Pert \cite{wang2021pert} integrates an erasing block that is repeatedly applied, combining a text localization module with a background reconstruction module to iteratively refine results. PSSTRNet \cite{lyu2022psstrnet} introduces a mask update module that incrementally refines text segmentation maps, employing attention mechanisms guided by the output of the previous iteration. Bian et al. \cite{bian2022scene} propose a comprehensive four-stage model, beginning with region-level mask processing through a detection-then-inpainting network. This framework then generates a stroke-level mask and an initial coarse result, which are further enhanced by a follow-up network using both masks. Concurrently, PEN \cite{du2023progressive} details an intermediate self-supervision approach based on the similarity of text stroke masks from augmented image versions, demonstrating enhanced performance in real-world scenarios through pretraining on synthetic data.

% \begin{figure*}[t]
%  \setlength{\abovecaptionskip}{0cm} %调整图片标题与图距离
% \begin{center}
% \includegraphics[width=1\textwidth]{fig/ste_pipeline.pdf}
% \hfill
% \end{center}
% \vspace{-10pt}

% \caption{\textbf{Scene text editing} methods mainly utilize the style feature to facilitate divide-and-conquer explicit transfer, implicit transfer, or inpainting-based framework. Text stoke also is considered by template or text prompt representation. To improve readability, text semantics is leveraged by semantic supervision.   }
% \vspace{-15pt}
% \label{fig:ste_pipeline}
% \end{figure*}

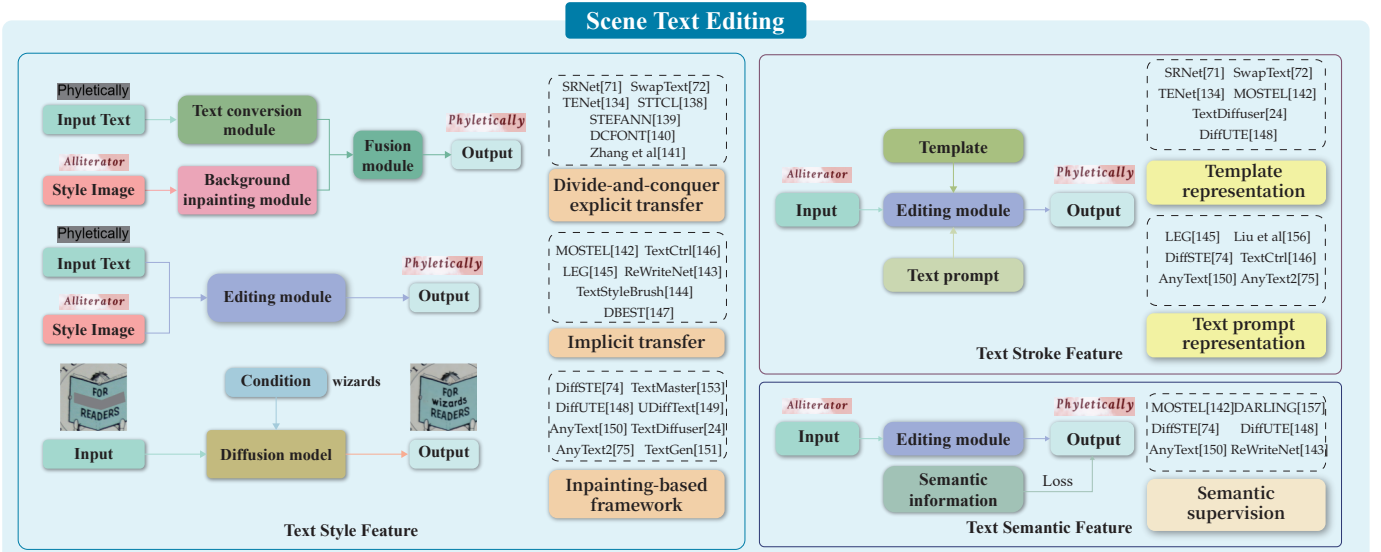
\begin{figure*}[t]
\setlength{\abovecaptionskip}{0cm}
\begin{center}
\begin{overpic}[width=\textwidth]{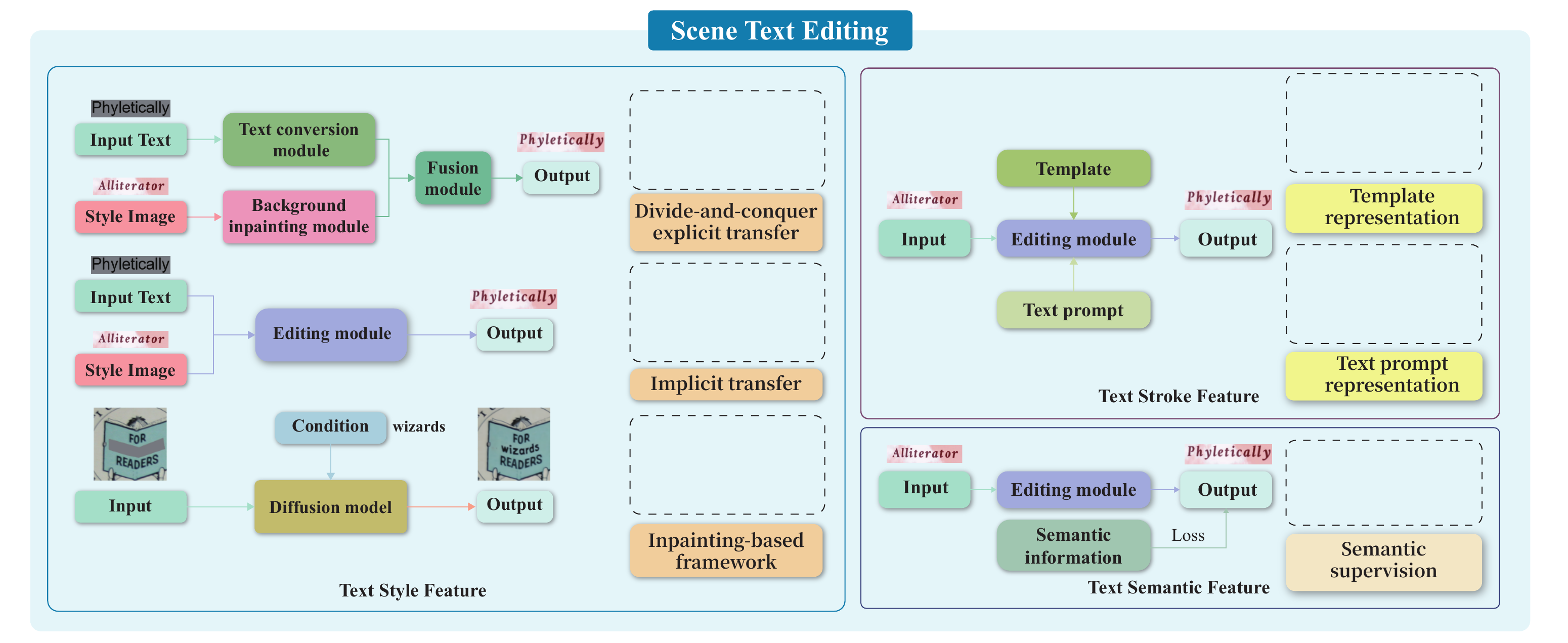}
\put(41,35){\hyperlink{cite.wu2019editing}{\textcolor{Black}{\tiny SRNet[71]}}}
\put(46,35){\hyperlink{cite.yang2020swaptext}{\textcolor{Black}{\tiny SwapText[72]}}}
\put(41,33.8){\hyperlink{cite.zhao2021deep}{\textcolor{Black}{\tiny TENet[134]}}}
\put(46.5,33.8){\hyperlink{cite.zhang2021scene}{\textcolor{Black}{\tiny STTCL[138]}}}
\put(43,32.6){\hyperlink{cite.roy2020stefann}{\textcolor{Black}{\tiny STEFANN[139]}}}
\put(43,31.4){\hyperlink{cite.roy2020stefann}{\textcolor{Black}{\tiny DCFONT[140]}}}
\put(43,30.2){\hyperlink{cite.roy2020stefann}{\textcolor{Black}{\tiny Zhang et al[141]}}}
\put(40.5,23){\hyperlink{cite.qu2023exploring}{\textcolor{Black}{\tiny MOSTEL[142]}}}
\put(47,23){\hyperlink{cite.zeng2024textctrl}{\textcolor{Black}{\tiny TextCtrl[146]}}}
\put(41,21.5){\hyperlink{cite.wang2023letter}{\textcolor{Black}{\tiny LEG[145]}}}
\put(45.5,21.5){\hyperlink{cite.lee2021rewritenet}{\textcolor{Black}{\tiny ReWriteNet[143]}}}
\put(42,20){\hyperlink{cite.krishnan2023textstylebrush}{\textcolor{Black}{\tiny TextStyleBrush[144]}}}
\put(44,18.5){\hyperlink{cite.santoso2023manipulating}{\textcolor{Black}{\tiny DBEST[147]}}}
\put(40.5,13){\hyperlink{cite.ji2023improving}{\textcolor{Black}{\tiny DiffSTE[74]}}}
\put(46,13){\hyperlink{cite.wang2024textmaster}{\textcolor{Black}{\tiny TextMaster[153]}}}
\put(40.5,11.5){\hyperlink{cite.chen2023diffute}{\textcolor{Black}{\tiny DiffUTE[148]}}}
\put(46.5,11.5){\hyperlink{cite.zhao2023udifftext}{\textcolor{Black}{\tiny UDiffText[149]}}}
\put(40,10){\hyperlink{cite.tuo2023anytext}{\textcolor{Black}{\tiny AnyText[150]}}}
\put(40.5,8.5){\hyperlink{cite.tuo2024anytext2}{\textcolor{Black}{\tiny AnyText2[75]}}}
\put(46,10){\hyperlink{cite.chen2023textdiffuser}{\textcolor{Black}{\tiny TextDiffuser[24]}}}
\put(47,8.5){\hyperlink{cite.zhang2024control}{\textcolor{Black}{\tiny TextGen[151]}}}
\put(85,36){\hyperlink{cite.wu2019editing}{\textcolor{Black}{\tiny SRNet[71]}}}
\put(90,36){\hyperlink{cite.yang2020swaptext}{\textcolor{Black}{\tiny SwapText[72]}}}
\put(84.5,34.5){\hyperlink{cite.zhao2021deep}{\textcolor{Black}{\tiny TENet[134]}}}
\put(90,34.5){\hyperlink{cite.qu2023exploring}{\textcolor{Black}{\tiny MOSTEL[142]}}}
\put(87,33){\hyperlink{cite.chen2023textdiffuser}{\textcolor{Black}{\tiny TextDiffuser[24]}}}
\put(87.5,31.5){\hyperlink{cite.chen2023diffute}{\textcolor{Black}{\tiny DiffUTE[148]}}}
\put(85,24){\hyperlink{cite.wang2023letter}{\textcolor{Black}{\tiny LEG[145]}}}
\put(90,24){\hyperlink{cite.liu2022character}{\textcolor{Black}{\tiny Liu et al[156]}}}
\put(85,22.5){\hyperlink{cite.ji2023improving}{\textcolor{Black}{\tiny DiffSTE[74]}}}
\put(90.5,22.5){\hyperlink{cite.zeng2024textctrl}{\textcolor{Black}{\tiny TextCtrl[146]}}}
\put(84.5,21){\hyperlink{cite.tuo2023anytext}{\textcolor{Black}{\tiny AnyText[150]}}}
\put(90.5,21){\hyperlink{cite.tuo2024anytext2}{\textcolor{Black}{\tiny AnyText2[75]}}}
\put(84,11.5){\hyperlink{cite.qu2023exploring}{\textcolor{Black}{\tiny MOSTEL[142]}}}
\put(84,10){\hyperlink{cite.ji2023improving}{\textcolor{Black}{\tiny DiffSTE[74]}}}
\put(89.8,8.5){\hyperlink{cite.lee2021rewritenet}{\textcolor{Black}{\tiny ReWriteNet[143]}}}
\put(90.5,10){\hyperlink{cite.chen2023diffute}{\textcolor{Black}{\tiny DiffUTE[148]}}}
\put(83.8,8.5){\hyperlink{cite.tuo2023anytext}{\textcolor{Black}{\tiny AnyText[150]}}}
\put(90,11.5){\hyperlink{cite.zhang2024choose}{\textcolor{Black}{\tiny DARLING[157]}}}
\end{overpic}
\hfill
\end{center}
\vspace{-10pt}
\caption{\textbf{Scene text editing} methods mainly utilize the style feature to facilitate divide-and-conquer explicit transfer, implicit transfer, or inpainting-based framework. Text stoke is also considered by template or text prompt representation. To improve readability, text semantics is leveraged by semantic supervision.   }
\vspace{-15pt}
\label{fig:ste_pipeline}
\end{figure*}

\subsection{Scene Text Editing}
Scene text is a composite visual element that generally appears on billboards and consists of various basic elements. Characters in the text share a harmonious decorative grammar, and their collective aesthetic speaks a unified visual language. Scene text editing aims to capture and distill the essence of text style to either adjust certain attributes or modify text contents. The overview of scene text editing methods is illustrated in Figure \ref{fig:ste_pipeline}.

\subsubsection{Text Style Feature}
Text style features encompass a range of inherent attributes such as font type, color, size, and space layout. These styles can be either implicitly learned in a latent space using style transfer networks or explicitly defined through fixed attributes.

\textbf{Divide-and-Conquer Explicit Transfer.} The text conversion module, first introduced by SRNet \cite{wu2019editing}, is employed to transfer the foreground text style from the source image to the target template image. After generating the modified target text, the network fuses it with the in-painted background from the source image to produce the final edited images. To adapt to text-style learning, skeleton-guided learning mechanisms are leveraged for fine-grained supervision. 

This strategy has been inherited and improved by several subsequent works \cite{yang2020swaptext,zhao2021deep,radford2021learning,yang2023self,das2023fast}. SwapText \cite{yang2020swaptext} integrates a shape transformation network for text shape control. TENet \cite{zhao2021deep} adopts a hard-coded component
\cite{radford2021learning} for text skeleton extraction of Chinese characters. STTCL \cite{zhang2021scene} further extends the network for cross-language scenery.

The design of the explicit transfer networks can be traced back to bitmap-based typeface learning \cite{roy2020stefann,jiang2017dcfont,zhang2018separating}, where image-to-image (I2I) translation models were initially applied for typeface generation. Nevertheless, the text style includes and surpasses pure typeface, making transfer learning more difficult. Besides, unsatisfactory transfer results can cause error accumulation in the subsequent fusion process. 

\textbf{Implicit Transfer.} Implicit text style transfer differs from previous methods by partially or completely discarding the intermediate image decoding process to avoid error accumulation. The partial methods \cite{qu2023exploring,das2023fast,su2023scene} integrate the conversion process or the in-painting process into the final fusion, while the complete methods \cite{lee2021rewritenet,krishnan2023textstylebrush,wang2023letter,zeng2024textctrl} wrap text editing as a conditional generation task based on the input of the source image and target text. 

While the editing pipeline has been simplified, these works improve their performance through various aspects. MOSTEL \cite{qu2023exploring} proposes semi-supervised learning from unpaired scene text data, utilizing augmented style reference and recognition loss to enhance training. TextStyleBrush \cite{krishnan2023textstylebrush} adopts a discriminator-based adversarial loss along with cyclic reconstruction to improve style consistency in the generated text.  DBEST \cite{santoso2023manipulating} and LEG \cite{wang2023letter} further leverage diffusion models while TextCtrl \cite{zeng2024textctrl} designs a style disentanglement pre-training strategy for attribute capture. These auxiliary designs on training strategies, network designs, and additional supervision further facilitated the exploration of editing.

\textbf{Inpainting-based Framework.} Powered by pre-trained diffusion models, the inpainting strategy is also leveraged for scene text editing. DiffSTE \cite{ji2023improving} improves pre-trained diffusion models with a dual encoder design, incorporating a character encoder for render accuracy and an instruction encoder for style control. DiffUTE \cite{chen2023diffute} replaces the CLIP text encoder with an OCR-based image encoder, improving text-style fidelity. Moreover, TextDiffuser \cite{chen2023textdiffuser} and UDiffText \cite{zhao2023udifftext} leverage character segmentation masks as conditioning inputs or supervised labels, respectively. AnyText \cite{tuo2023anytext} and TextGen \cite{zhang2024control} adopt a universal framework to resolve editing in multiple languages based on the prevalent ControlNet \cite{zhang2023adding}. Additionally, TextMaster \cite{wang2024textmaster} employs adaptive spacing and mask control to enhance the learning of text layout capabilities.

Although inpainting-based methods naturally enable self-supervised learning on extensive real-world text images, most of current methods neglect the consistency of text style. The idea of style decoupling process on font style and conditioning the style with an IP-Adapter \cite{ye2023ip} in TextMaster \cite{wang2024textmaster}, and attributes customization through text embedding in DiffSTE \cite{ji2023improving} and AnyText2 \cite{tuo2024anytext2} can be seen as a small step of experimentation. However, further exploration is needed to explore more effective style transfer methods.

\subsubsection{Text Stroke Feature}
Text stroke feature plays an essential role in scene text editing serving as the condition for glyph rendering. The design of text stroke in editing mainly focuses on two aspects, namely template representation and text prompt representation.

% \begin{figure*}[t]
%  \setlength{\abovecaptionskip}{0cm} %调整图片标题与图距离
% \begin{center}
% \includegraphics[width=1\textwidth]{fig/stg_pipeline1.pdf}
% \hfill
% \end{center}
% \vspace{-10pt}
% \caption{\textbf{Scene text generation} methods mainly utilize the structure feature to realize geometry-aware text synthesis and generative layout control. To generate images resemble real-world data,  the style feature is integrated into adversarial style learning. To render high-fidelity texts, the stroke feature is as glyph template prior or glyph embedding for condition input in diffusion models. }
% \vspace{-15pt}
% \label{fig:stg_pipeline}
% \end{figure*}

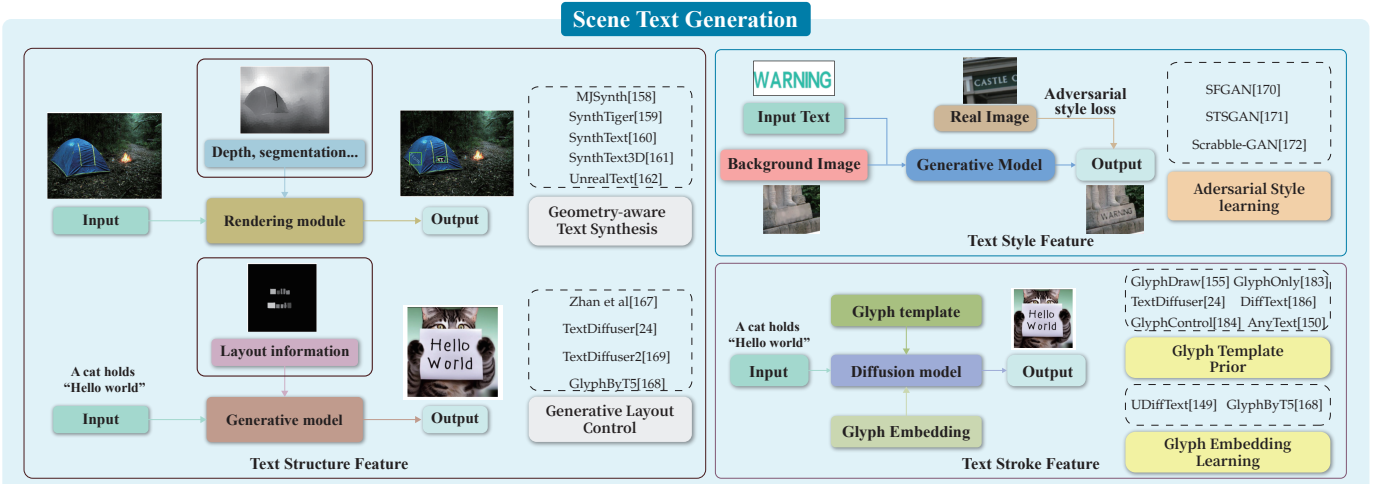
\begin{figure*}[t]
\setlength{\abovecaptionskip}{0cm}
\begin{center}
\begin{overpic}[width=\textwidth]{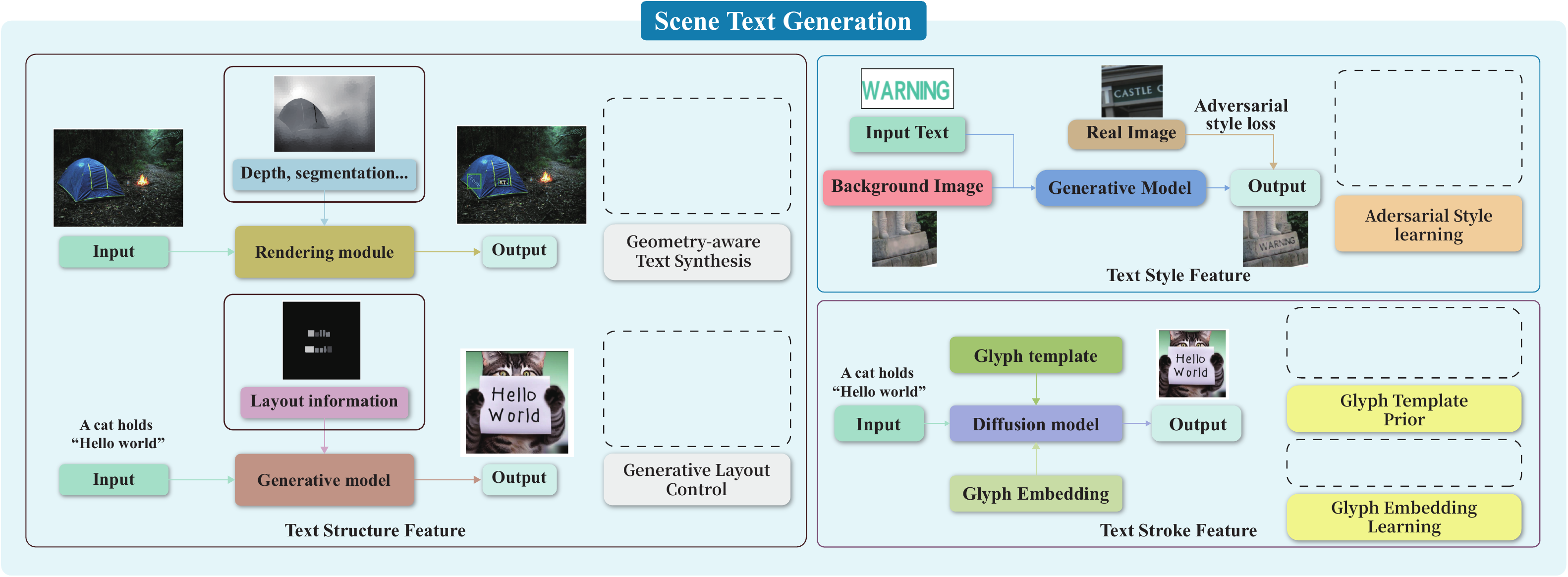}
\put(42,29.5){\hyperlink{cite.jaderberg2014synthetic}{\textcolor{Black}{\tiny MJSynth[158]}}}
\put(41.5,28){\hyperlink{cite.yim2021synthtiger}{\textcolor{Black}{\tiny SynthTiger[159]}}}
\put(41.5,26.5){\hyperlink{cite.gupta2016synthetic}{\textcolor{Black}{\tiny SynthText[160]}}}
\put(41.5,25){\hyperlink{cite.liao2020synthtext3d}{\textcolor{Black}{\tiny SynthText3D[161]}}}
\put(41.5,23.5){\hyperlink{cite.long2020unrealtext}{\textcolor{Black}{\tiny UnrealText[162]}}}
\put(41.5,14.5){\hyperlink{cite.zhan2018verisimilar}{\textcolor{Black}{\tiny Zhan et al[167]}}}
\put(41,12.5){\hyperlink{cite.chen2023textdiffuser}{\textcolor{Black}{\tiny TextDiffuser[24]}}}
\put(41,10.5){\hyperlink{cite.chen2024textdiffuser2}{\textcolor{Black}{\tiny TextDiffuser2[169]}}}
\put(41.5,8.5){\hyperlink{cite.liu2024glyph}{\textcolor{Black}{\tiny GlyphByT5[168]}}}
\put(88,30){\hyperlink{cite.zhan2019spatial}{\textcolor{Black}{\tiny SFGAN[170]}}}
\put(88,28){\hyperlink{cite.fang2019learning}{\textcolor{Black}{\tiny STSGAN[171]}}}
\put(87,26){\hyperlink{cite.fogel2020scrabblegan}{\textcolor{Black}{\tiny Scrabble-GAN[172]}}}
\put(82.5,16){\hyperlink{cite.ma2023glyphdraw}{\textcolor{Black}{\tiny GlyphDraw[155]}}}
\put(90,16){\hyperlink{cite.li2024first}{\textcolor{Black}{\tiny GlyphOnly[183]}}}
\put(82.5,14.5){\hyperlink{cite.chen2023textdiffuser}{\textcolor{Black}{\tiny TextDiffuser[24]}}}
\put(90.5,14.5){\hyperlink{cite.zhang2024brush}{\textcolor{Black}{\tiny DiffText[186]}}}
\put(82.5,13){\hyperlink{cite.yang2023glyphcontrol}{\textcolor{Black}{\tiny GlyphControl[184]}}}
\put(91,13){\hyperlink{cite.tuo2023anytext}{\textcolor{Black}{\tiny AnyText[150]}}}
\put(82.5,7){\hyperlink{cite.zhao2023udifftext}{\textcolor{Black}{\tiny UDiffText[149]}}}
\put(89.5,7){\hyperlink{cite.liu2024glyph}{\textcolor{Black}{\tiny GlyphByT5[168]}}}
\end{overpic}
\hfill
\end{center}
\vspace{-10pt}
\caption{\textbf{Scene text generation} methods mainly utilize the structure feature to realize geometry-aware text synthesis and generative layout control. To generate images resemble real-world data,  the style feature is integrated into adversarial style learning. To render high-fidelity texts, the stroke feature is as glyph template prior or glyph embedding for condition input in diffusion models. }
% \vspace{-15pt}
\label{fig:stg_pipeline}
\end{figure*}

\textbf{Template Representation.} Early GANs-based editing methods \cite{wu2019editing,yang2020swaptext,zhao2021deep,radford2021learning,qu2023exploring} rely on a conversion network, which adopts template text image as model's input. The template representation provides explicit stroke guidance to prevent randomness in result characters. Inpainting-based diffusion methods \cite{chen2023textdiffuser,ma2023glyphdraw,chen2023diffute} also leverage the image of text template in the model design, either concatenating it with latent attributes or encoding it to serve as a condition in the cross-attention.

\textbf{Text Prompt Representation.} With the incorporation of fine-grained character-level text encoders, text stroke can also be represented using prompt representations \cite{wang2023letter,zeng2024textctrl}. Addressing the nuances of text embedding representation, Liu et al. \cite{liu2022character} highlight the impact of overlooking character-level input features on the fidelity of visual text. Their study demonstrates that a shift from character-blind input tokens to character-aware tokens markedly improves the spelling precision of visual text. To further align with the text-guided diffusion models, recent methods \cite{tuo2023anytext,ji2023improving} adopt multi-encoders for text prompt, enabling fine-grained representation for rendered text. Further exploration on modality alignment is expected to further enhance the effectiveness of prompt representations.

\subsubsection{Text Semantic Feature}
Text sequences encompass more than just a series of characters; they also carry rich semantic information that can guide the restoration or modification of text images. To enhance the learning of these semantic features, many methods incorporate an auxiliary text recognition loss.

\textbf{Text Recognition Supervision.} The text recognition loss is widely used in text editing training to provide auxiliary supervision for ensuring rendering accuracy. Early GAN-based methods \cite{qu2023exploring,lee2021rewritenet} leverage the recognition loss in self-supervised learning since it does not require paired data. Diffusion-based methods \cite{tuo2023anytext,chen2023diffute,ji2023improving} further adopt the semantic loss for the decoded image. In addition, recent work \cite{zhang2024choose} integrates text editing and text recognition training in a single module for better visual and content representation disentanglement, which further inspires the exploration of how semantic information assists in editing.

\subsection{Scene Text Generation}
Scene text generation aims to produce natural, visually harmonious, and contextually coherent text within images. Existing frameworks leverage text structure, style, and semantic features to optimize text layout, font style, and readability (See Figure \ref{fig:stg_pipeline}).

% features are utilized for optimizing layout, while 

% textual images. This technology plays a pivotal role not only in generating datasets for scene text detection and recognition but also finds broad applications in fields such as poster design and storefront signage. More specifically, scene text generation can be categorized into two primary paradigms: Scene Text Rendering and Whole Scene Text Image Generation.

\subsubsection{Text Structure Feature}

Structural features capture rich geometric knowledge in scene texts, which can be leveraged to better imitate real-world text distributions. Furthermore, enhancing the relationship between text structure and local backgrounds through layout control is crucial for generating visually plausible texts.

\textbf{Geometry-aware Text Synthesis.} A key challenge in scene text synthesis is generating text that follows real-world geometric distributions. To address this, various methods focus on designing generators that overlay synthetic text onto backgrounds while ensuring realistic spatial alignment. MJSynth \cite{jaderberg2014synthetic} incorporates border/shadow rendering and perspective distortion into its blending pipeline, enabling the synthesis of curved text. SynthTiger \cite{yim2021synthtiger} extends this framework by introducing elastic distortion in text shape selection, further enhancing text variation. Rather than generating word box images, SynthText \cite{gupta2016synthetic} focuses on rendering text directly within natural images. To achieve this, it considers local scene geometry relevant to text placement, including depth and segmentation information. Building upon this framework, SynthText3D \cite{liao2020synthtext3d} and UnrealText \cite{long2020unrealtext} introduce 3D virtual scene rendering, integrating text instances seamlessly into realistic environments. These methods enable more complex perspective transformations through a 3D rendering module, significantly improving text realism in synthetic data.

\textbf{Generative Layout Control.} Generative foundation models \cite{dhariwal2021diffusion, xia2024diffusion,he2025diffusion,zhan2023multimodal} are widely used in scene text generation. Many studies focus on designing layout control mechanisms to optimize the relationship between text structure and local backgrounds. Zhan et al. \cite{zhan2018verisimilar} introduce semantic coherence and saliency maps to ensure text is embedded in semantically meaningful regions. Within the diffusion model family, TextDiffuser \cite{chen2023textdiffuser} incorporates a layout transformer capable of learning character positions and layout structures. GlyphByT5 \cite{liu2024glyph} employs a region-level multi-head encoder fusion mechanism to facilitate multi-line text layout generation. TextDiffuser2 \cite{chen2024textdiffuser2} leverages large language models to convert user instructions into layout positions, enabling more flexible interactions. Building upon the effective Diffusion Transformer (DIT) architecture, TextFlux \cite{xie2025textflux} achieves the multilingual Scene Text Synthesis by dealing with the glyph and the text mask regions  through an efficient concatenation scheme. TextCrafter \cite{du2025textcrafter} strength the relationship between visual text and its corresponding carrier by Instance Fusion method, thereby rendering texts on user intended objects. BizGen \cite{peng2025bizgen} try to address article-level text synthesizing by designing a Layout Guided Cross-Attention method, ensuring texts are rendered in each sub-regions.

\subsubsection{Text Style Feature}
Since GANs have demonstrated strong capabilities in style transfer, many studies have adapted them for scene text synthesis by mimicking real-world text styles, including font, color, and texture.

\textbf{Adversarial Style Learning.} SFGAN \cite{zhan2019spatial} introduces an appearance synthesizer that adjusts the color, brightness, and style of foreground objects, enabling seamless text and background integration. STS-GAN \cite{fang2019learning} addresses the challenge of generating characters with a consistent style. To mitigate style variations among characters, it introduces a novel adversarial style loss based on variance minimization. Additionally, Scrabble-GAN \cite{fogel2020scrabblegan} employs a semi-supervised approach to generate handwritten text images with diverse styles and vocabularies. Its architecture consists of individual character generators, a style-controlling discriminator, and a text recognizer, ensuring that the generated text remains realistic and legible.

Besides, the concurrent work FonTS \cite{shi2024fonts} trys to enhance controllability over typography and style in text rendering process by proposing typography control fine-tuning method. Moreover, Style Control Adapters are designed to douple content and style learning in the training process.

\begin{table*}[t]
\renewcommand{\arraystretch}{1}
\centering
\caption{Statistics of visual text processing benchmarks, including data size, language, source (``Syn" denotes synthetic and "Real" indicates real-world), type (scene, document, or designed poster), scope (original images or cropped regions), method (human-annotated or model-generated). } 
\label{tab:dataall}
\vspace{-10pt}
\resizebox{1\linewidth}{!}{
\begin{tabular}{ccccccccc}
\toprule
Task & Dataset      & Year & Size   & Language          & Source & Type           & Scope  & Method \\ \midrule
\multirow{2}{*}{Text Image Super-Resolution} & TextZoom \cite{Wang2020Scene}        & 2020 & 8746 & English      & Real       & Scene          & Region  & Human         \\ 
& Real-CE\cite{ma2023benchmark}        & 2023 & 300 & English + Chinese & Real       & Scene          & Region  & Human         \\ 
\midrule
\multirow{5}{*}{Document Image Dewarping}& DocUNet\cite{Ma2018DocUNetDI}       & 2018  & 130           & Multilingual           & Real      & Document      & Whole  & Human         \\
                                        %& DRIC\cite{Li2019DocumentRA}          & 2019  & 1,300         & English           & Syn      & Document      & Whole  & Human  \\
                                        %& Doc3D\cite{Das2019DewarpNetSD}         & 2019  & 100,000       & English           & Syn      & Document      & Whole  & Human  \\
                                        & DIR300\cite{Feng2022GeometricRL}        & 2022  & 300           & English           & Real      & Document      & Whole  & Human  \\
                                        %& WarpDoc\cite{Xue2022FourierDR}       & 2022  & 1,020         & Englsih           & Real      & Document      & Whole  & Human  \\ \midrule
                                        & DocReal\cite{yu2024docreal} & 2023 & 200 & Chinese & Real & Document & Whole & Human \\
                                        & UVDoc\cite{verhoeven2023uvdoc} & 2023 & 50 & English & Syn & Document & Whole & Human \\
                                        \midrule
\multirow{5}{*}{Text Image Enhancement} & Kligler et al. \cite{Kligler2018DocumentEU} & 2018 & 300 & Multilingual & Real & Document & Whole & Human\\
& DIBCO'18 \cite{pratikakis2018icfhr} & 2018 & 10 & English & Real & Document & Whole & Human\\
& OSR \cite{wang2020local} & 2020 & 237 & English & Real & Document & Region & Human \\
& RealDAE\cite{zhang2023appearance} & 2023 & 150 & Multilingual & Real & Document & Whole + Region & Human\\
& RDD \cite{zhang2023document} & 2023 & 545 & Multilingual & Real & Document & Region & Human \\
\midrule
% \multirow{4}{*}{Scene Text Segmentation} & COCO\_TS        & 2019 & 14,690 & English      & Real       & Scene          & Whole  & Model         \\
%      & MLT\_TS      & 2019 & 6,896  & English           & Real   & Scene          & Whole  & Model  \\
%      & TextSeg      & 2021 & 4,024  & English           & Real   & Scene + Design & Whole  & Human  \\
%      & BTS          & 2022 & 14,250 & Englsih + Chinese & Real   & Scene          & Whole  & Human  \\ \hline
\multirow{5}{*}{Scene Text Removal}      & SCUT-Syn \cite{zhang2019ensnet}        & 2019 & 800  & English      & Syn        & Scene          & Whole  & Model         \\
     & SCUT-EnsText \cite{liu2020erasenet} & 2020 & 813  & English           & Real   & Scene          & Whole  & Human  \\
                                       
     & PosterErase  \cite{jiang2022self} & 2022 & 400 & Chinese           & Real   & Design         & Whole  & Human  \\
     & Flickr-ST \cite{lyu2023fetnet}   & 2023 & 800  & English           & Real   & Scene          & Whole  & Human  \\ \midrule
\multirow{3}{*}{Scene Text Editing}      & SynthText-Based \cite{gupta2016synthetic} & 2019 & -      & English      & Syn        & Scene          & Region & Model         \\
 & Tamper \cite{qu2023exploring}     & 2023 & 159,725    & English           & Real + Syn   & Scene          & Region & Human  \\ 
  & ScenePair \cite{zeng2024textctrl}     & 2024 & 1,280   & English           & Real + Syn   & Scene          & Region & Human  \\ 
 \midrule

\multirow{4}{*}{Scene Text Generation}   & MARIO-Eval \cite{chen2023textdiffuser}            & 2023 & 5,000    & English      & Real       & Scene  & Whole + Region  & Human + Model         \\
                                         & DrawTextExt \cite{ma2023glyphdraw}     & 2023 & -   & English + Chinese      & Real       & Scene + Design & Whole  & Model         \\
 & AnyText \cite{tuo2023anytext}            & 2023 & 1,000    & English + Chinese       & Real       & Scene + Design & Whole  & Model       \\
  & VisualParagraphy  \cite{liu2024glyph}            & 2024 & 1,000    & English        & Syn       & Design & Whole  & Model       \\
                                    \bottomrule
\end{tabular}}
% \vspace{-5pt}
\end{table*}

\subsubsection{Text Stroke Feature}
Fine-tuning diffusion models for scene text rendering has become the mainstream approach due to their exceptional generative capabilities. Within this paradigm, many studies have incorporated stroke knowledge to achieve accurate and legible text rendering.

\textbf{Glyph Template Prior.}  By integrating cross-attention layers into the model architecture, Latent Diffusion~\cite{rombach2022high} achieves a powerful and flexible generator conditioned on text, image, and audio. A straightforward approach is to use a glyph template as a specific image condition to generate desired visual text. GlyphDraw \cite{ma2023glyphdraw} employs a pre-trained CLIP model to encode glyph images into embeddings and establishes a fusion module to aggregate text input and glyph embeddings as conditioning input. Building on this framework, TextDiffuser \cite{chen2023textdiffuser} and GlyphOnly \cite{li2024first} incorporate a character-aware loss to help the model focus more on text regions during the generation process. Inspired by ControlNet \cite{zhang2023adding}, which implements zero convolution to learn rich conditional representations such as edge, depth, and segmentation, GlyphControl \cite{yang2023glyphcontrol} and STGen \cite{luo2025beyond} introduce glyph-conditioned information without retraining the diffusion model, thereby preserving its internal generative capabilities. Similarly, DiffText \cite{zhang2024brush} proposes a training-free approach that leverages rendered sketch images as priors, enhancing the pre-trained Stable Diffusion model’s potential for multilingual text generation.  AnyText \cite{tuo2023anytext} and SceneVTG \cite{zhu2024visual} employ a pretrained text recognition model for glyph image encoding, coupled with a text perceptual loss to further improve the accuracy of text generation.

\textbf{Glyph Embedding Learning.} Although fixed glyph templates can produce legible text, rendering errors often arise due to misalignment between glyph features and the knowledge embedded in diffusion models. To address this issue, UDiffText \cite{zhao2023udifftext} replaces the original CLIP encoder with a lightweight character-level text encoder to provide more robust text embeddings. Similarly, GlyphByT5 \cite{liu2024glyph} introduces character-aware text encoders, trained to capture the rich information encoded within visual glyph representations extracted from a pre-trained image model. Specifically, during training, only the text encoders are updated, while all other components remain frozen, ensuring stability and efficient adaptation to glyph features. PosterMarker \cite{gao2025postermaker} proposes the TextRenderNet to obtain character-discriminative visual features, thus facilitating accurate text rendering.

\begin{table*}[t]
\scriptsize
\renewcommand{\arraystretch}{1}
		\centering
		 \caption{Text image super-resolution methods on TextZoom. Bold denotes the \textbf{best} result, and underline denotes the \underline{second-best} result.}
		\label{tab:TextZoom}
  \vspace{-10pt}
		\begin{tabular}{ccccccccc}
			\toprule
			\multirow{2}{*}{Methods}                & \multicolumn{4}{c}{Recognition Accuracy}                         & \multicolumn{4}{c}{Image Quality (PSNR/SSIM)} \\ 
			\cmidrule(r){2-5} \cmidrule(r){6-9}
			 & Easy    & Medium  & Hard   & \bf{Average} $\uparrow$   & Easy    & Medium  & Hard   & \bf{Average} $\uparrow$  \\ \midrule
LR                     &62.40\%    &42.70\%    &31.60\%    &46.58\%    &  - &  - & -  &-\\
\midrule
Bicubic                     &64.70\%    &42.40\%    &31.20\%    &47.20\%    &22.3500/0.7884   &18.9800/0.6254   &19.3900/0.6592   &20.3500/0.6961\\
TSRN \cite{Wang2020Scene}  &75.10\%    &56.30\%    &40.10\%	&58.30\%    &25.0700/0.8897	&18.8600/0.6676	&19.7100/0.7302	&21.4200/0.7690\\
TPGSR \cite{Ma2021TextPG}    &78.90\%    &62.70\%    &44.50\%	&62.80\%	&23.7300/0.8805	&18.6800/0.6738	&20.0600/0.7440	&20.9700/0.7719\\
TBSRN \cite{Chen2021SceneTT} &75.70\%	&59.90\%	&41.60\%	&60.10\%	&23.8200/0.8660	&19.1700/0.6533	&19.6800/0.7490	&20.9100/0.7603\\
PCAN \cite{Zhao2021SceneTI}  &77.50\%	&60.70\%	&43.10\%	&61.50\%	&24.5700/0.8830	&19.1400/0.6781	&20.2600/0.7475	&21.4900/0.7752\\
TG \cite{Chen2021TextGS}     &77.90\%	&60.20\%	&42.40\%	&61.30\%	&23.3400/0.8369	&\underline{19.6600/0.6499}	&19.9000/0.6986	&21.4000/0.7456\\
TATT \cite{Ma2022ATA}        &78.90\%	&63.40\%	&45.40\%	&63.60\%	&\underline{24.7200/0.9006}	&19.0200/0.6911	&\underline{20.3100/0.7703}	&\underline{21.5200/0.7930}\\
C3-STISR \cite{Zhao2022C3STISRST} &79.10\%	&63.30\%	&46.80\%	&64.10\%	&- &- &- &21.5100/0.7721\\
DPMN \cite{Zhu2023ImprovingST}&79.25\%	&\underline{64.07\%}	&45.20\%	&63.89\%	&- &- &-	&21.4900/0.7925\\
TSEPG \cite{Huang2023TextIS} &\underline{79.60\%}	&63.90\%	&\textbf{47.50\%}	&\underline{64.68\%}	&\textbf{25.3600/0.9053}	&\textbf{20.2600/0.6931}	&\textbf{20.5800/0.7782}	&\textbf{22.2500/0.7978}\\
LEMMA \cite{guo2023towards} &\textbf{81.10\%}	&\textbf{66.30\%}	&\underline{47.40\%}	&\textbf{66.00\%}	&- &- &-	&{20.9000/0.7792}\\

\midrule
HR  &94.20\%	&87.70\%	&76.20\%	&86.60\% & - & - & - & - \\

\bottomrule
		\end{tabular}
        \vspace{-10pt}
	\end{table*}

\section{Benchmark and Evaluation}
\label{section4}
% The swift advancement of visual text processing tasks and algorithms has been paralleled by a significant growth in datasets for training and evaluation. This section offers an overview of the prominent datasets, with key characteristics summarization in Table \ref{dataall} and  comprehensive review in the following discussion.

In this section, we first review existing benchmarks and evaluation metrics used in visual text processing. Then, we present a detailed construction of the proposed VTPBench and explain the detailed implementation of VTPScore. Finally, we discuss empirical results and analysis.

\subsection{Existing Benchmarks and Evaluation}
\label{sec:data}
Here we show mainstream benchmarks (see Table \ref{tab:dataall}) and evaluation metrics used in different visual text processing domains.

\textbf{Text Image Super-resolution.} TextZoom \cite{Wang2020Scene} is the first real-world dataset containing camera-captured low-resolution (LR)–high-resolution (HR) text image pairs with varying focal lengths. It provides image pairs, text labels, bounding box types, and original focal lengths. In addition to English text super-resolution, Real-CE \cite{ma2023benchmark} focuses on restoring structurally complex Chinese characters. Peak signal-to-noise ratio (PSNR) and structural similarity index measure (SSIM) are primarily used for image-level evaluation. Additionally, recognition accuracy is adopted to assess restored text readability. 

% TextZoom is one of the leading benchmarks in text image super-resolution (TISR). As shown in Table \ref{tab:TextZoom}, TSEPG emerges as the top-performing method in TISR, excelling in both recognition accuracy and image quality. However, it still falls significantly short of the theoretical upper bound, which is determined by high-resolution images.

\textbf{Document Image Dewarping.} DocUNet \cite{Ma2018DocUNetDI} consists of 65 paper documents captured by mobile cameras in two distorted shapes, resulting in 130 images in total, along with corresponding flat-scanned images as ground truth. The dataset includes various document types, such as receipts, letters, flyers, magazines, academic papers, and books. DIR300 \cite{Feng2022GeometricRL} features more complex backgrounds and diverse illumination conditions in the test set. Specifically, images are captured using different cellphones in various environments under multiple distortions, including curved, folded, flat, and heavily crumpled documents. Unlike DocUNet and DIR300, DocReal \cite{yu2024docreal} focuses on real-life Chinese document image scenarios. UVDoc \cite{verhoeven2023uvdoc} is also a photorealistic dataset which combines pseudo photorealistic document images with physically accurate 3D shapes and unwarping function annotations. Both image-level and OCR-level metrics are used for document image dewarping evaluation. For image-level metrics: Multi-scale structural similarity (MS-SSIM) \cite{1292216} extends SSIM across multiple scales via a Gaussian pyramid, assessing the global similarity between reconstructed and ground truth images. Local distortion (LD) \cite{4270276} calculates a dense SIFT flow from the reconstructed image to the ground truth scan, measuring the rectification quality of local details.
Aligned distortion (AD) \cite{Ma2022LearningFD} aligns the unwarped image with the scan image before evaluation and weighs the error based on gradient magnitude. For OCR-level metrics, edit distance (ED) and character error rate (CER) are computed on selected text-rich images within DocUNet to assess the recognition quality of the reconstructions. 

% DocUNet and DIR300 are the two most extensively used datasets in this domain. Quantitative results in Table \ref{tab:docunet} illustrate the progress of DID methods. Notably, the most advanced solution to date, as proposed by \cite{Li2023LayoutAwareSD}, incorporates both global and local features while utilizing more realistic datasets than previous methods. This approach achieves 0.526 in MS-SSIM and 6.72 in LD, marking a significant improvement over earlier DID techniques.

\textbf{Text Image Enhancement.} Text image enhancement involves distinct benchmarks for various enhancement tasks, including deblurring, deshadowing, illumination enhancement, binarization, and so on. The Text Deblurring Dataset \cite{hradivs2015convolutional} (TDD) stands as the most frequently employed dataset in the realm of deblurring, with each image presented as a cropped 300$\times$300 patch.
$\times$ 300 pixel patch. For deshadowing, Kligler et al. \cite{Kligler2018DocumentEU} construct a benchmark comprising 300 high-resolution images, including both handwritten and printed documents. OSR \cite{wang2020local} includes 237 document images of size 960 $\times$ 544, which are captured indoors. For illumination enhancement, RealDAE \cite{zhang2023appearance} is the first dataset that targets multiple degradations in the wild, which contains 600 real-world degraded document images that are carefully annotated with pixelwise alignment. From 2009 to 2019, DIBCO competition releases annual benchmarks of 10 or 20 document images for binarization task. Among these, DIBCO'18 \cite{pratikakis2018icfhr} is the most frequently used. PSNR and SSIM are widely used to evaluate the effectiveness of deblurring, deshadowing, and illumination enhancement.  For the binarization task, PSNR, F-measure (FM), and pseudo F-measure (pFM) are used as evaluation metrics. FM and pFM combine precision and recall to provide a better evaluation of the model's overall performance.

\textbf{Scene Text Removal.} SCUT-Syn \cite{zhang2019ensnet} is the first scene text removal benchmark, which utilizes text synthetic techniques to get image pairs. To bridge the gap between synthetic data and real-world images, SCUT-EnsText \cite{liu2020erasenet} is carefully designed, where each image is meticulously annotated to provide visually coherent erasure targets, with human-assisted editing using Adobe Photoshop. For Flickr-ST \cite{lyu2023fetnet}, this dataset offers comprehensive annotations, including text-removed images, pixel-level text masks, character instance segmentation labels, character category labels, and character-level bounding boxes. STR evaluation metrics include detection-eval and image-eval. Detection-eval focuses on the thoroughness of text region removal, using an auxiliary text detector to gather detection results post text removal and evaluates the precision, recall, and F-score. Image-Eval including the following aspects: (i) mean squared error (MSE); (ii) PSNR; (iii) SSIM; (iv) AGE, which calculates the average of the graylevel absolute difference between the ground truth and the computed background image; (v) pEPs, which calculate the percentage of error pixels; and (vi) pCEPS, which calculates the percentage of clustered error pixels.

% We report STR results on SCUT-Syn and SCUT-EnsText, as shown in Table \ref{tab:enstext}. 
%  On the SCUT-Syn dataset, MBE outperforms other methods in PSNR and SSIM due to its ensemble strategy. In contrast, ViTEraser achieves the best performance on SCUT-EnsText across most metrics. This is primarily because ViTEraser employs a self-training scheme during pre-training, allowing it to learn more effectively from real-world data.

\textbf{Scene Text Editing.} Early scene text editing (STE) methods relied on synthetic data for evaluation. Tamper \cite{qu2023exploring} is introduced as a composite dataset combining multiple scene text datasets to assess editing accuracy. ScenePair \cite{zeng2024textctrl}, the first real-world scene text editing benchmark, provides the source text image, the target text image, respective text labels, quadrangle locations in the full-size image, and the original full-size image. Analogous to scene text removal, image-eval metrics, including MSE, PSNR, SSIM, and Fréchet Inception Distance (FID) \cite{fid}, are used to assess the style similarity between edited images and ground truth. Additionally, text recognition accuracy measures text fidelity to the target text.

% We present the results on ScenePair and Tamper-Scene, as shown in Table \ref{tab:tamper}. Notably, TextCtrl demonstrates superior performance compared to other methods.

% It is critical to recognize that many scene text editing approaches predominantly utilize various synthetic datasets during their training and testing phases, leading to potential biases in performance evaluations. To promote a more balanced and fair assessment, our results are exclusively derived from the benchmark set by Qu et al. \cite{qu2023exploring}, as detailed in Table \ref{tamper}. In this particular dataset, VTNet \cite{susladkar2023towards} demonstrates superior performance compared to other methods.

\begin{table*}[]
\scriptsize
\renewcommand{\arraystretch}{1}
		\centering
		\caption{Document image dewarping  performance comparison on DocUNet and DIR300.  $\ast$ indicates experimental results  from the original paper of each method, with different OCR engine utilized. $\dagger$ indicates experimental results reported from \cite{gupta2021layouttransformer}, which use PyTesseract v0.3.9 for OCR testing.}
		\label{tab:docunet}
  \vspace{-10pt}

		\begin{tabular}{ccccccccccc}
			\toprule
			\multirow{2}{*}{Methods}                & \multicolumn{5}{c}{DocUNet$\ast$}                         & \multicolumn{5}{c}{DIR300$\dagger$} \\ 
			\cmidrule(r){2-6} \cmidrule(r){7-11}
			& MS-SSIM $\uparrow$  & LD  $\downarrow$  & AD $\downarrow$    & ED $\downarrow$    & CER (\%)$\downarrow$  & MS-SSIM $\uparrow$  & LD  $\downarrow$  & AD $\downarrow$    & ED $\downarrow$    & CER (\%) $\downarrow$      \\ \midrule
			DocUNet \cite{Ma2018DocUNetDI}        &0.4100	&14.08        &-      &-                      &-          &-              &- &-&-&- \\   
			DewarpNet \cite{Das2019DewarpNetSD}   &0.4735	&8.95      &0.426  &1114.4 	      &26.92 &0.4921 &13.94 &0.331 &1059.57 &35.57  \\ 
   DFCN \cite{Xie2020DewarpingDI}        &0.4361	&8.50	      &0.434      &-                      &-          &0.5035 & 9.75 & 0.331 &1939.48 &50.99         \\
AGUN \cite{Liu2020GeometricRO}        &0.4491	&12.06        &-      &-                      &-          &-              &-  &-&-&-\\  
  Piece-Wise \cite{Das2021EndtoendPU}    &0.4879	&9.23         &0.468  &- &30.01 &-&-&-&-&- \\
   DWCP \cite{Xie2022DocumentDW}         &0.4769	&9.03	      &0.453      &-                      &-          &0.5524 &10.95 &0.357 &2084.97 &54.10            \\
   DocTr \cite{Feng2021DocTrDI}          &0.4970	&8.38	      &0.396 &\underline{576.4} &20.00 &0.6160 &7.21 &0.254 &699.63 &22.37 \\
   DocScanner \cite{Feng2021DocScannerRD}&\underline{0.5178}	&\underline{7.45}	      &\underline{0.334}	&632.3 &\textbf{16.48} &-&-&-&-&- \\
   PaperEdge \cite{Ma2022LearningFD}     &0.4700	&8.50	      &0.392 &1010.0 &22.10 &0.5836 &8.00 &0.255 &\textbf{508.73} &\underline{20.69} \\
   Marior \cite{Zhang2022MariorMR}       &0.4733	&8.08	      &0.403 &- &18.35 &-&-&-&-&- \\
   RDGR \cite{Jiang2022RevisitingDI} &0.4922	&9.36	  &0.461 &896.5 &20.68 &-&-&-&-&- \\
FDR \cite{Xue2022FourierDR}           &0.5000	&9.43	      &- &- &16.96 &-&-&-&-&- \\
DocGeoNet \cite{Feng2022GeometricRL}     &0.5040	&7.71 &0.380 &713.9 &18.21 &\underline{0.6380} &\underline{6.40} &\underline{0.242} &664.96 &21.89 \\ 
DocTr++ \cite{Feng2023DeepUD}         &0.5100	&7.52	      &-  &\textbf{447.5} &\underline{16.95} &- &-&-&-&-\\
Li et al. \cite{Li2023LayoutAwareSD}  &\textbf{0.5260}	&\textbf{6.72}	&\textbf{0.300} &695.0 &17.50 & \textbf{0.6518} &\textbf{5.70} &\textbf{0.195} &\underline{511.13} &\textbf{18.91} \\

\bottomrule
		\end{tabular}
\vspace{-10pt}
	\end{table*}

\begin{table*}[]
\scriptsize
\renewcommand{\arraystretch}{1}
		\centering
		\caption{Scene text removal performance comparison on SCUT-EnsText and SCUT-Syn.}
		\label{tab:enstext}
  \vspace{-10pt}
		\begin{tabular}{ccccccccccc}
			\toprule
			\multirow{2}{*}{Methods}                & \multicolumn{7}{c}{SCUT-EnsText}                         & \multicolumn{3}{c}{SCUT-Syn} \\ 
			\cmidrule(r){2-8} \cmidrule(r){9-11}
			& PSNR $\uparrow$  & SSIM (\%) $\uparrow$  & MSE $\downarrow$    & AGE $\downarrow$    & pEPs $\downarrow$   & pCEPs $\downarrow$  & F $\downarrow$    & PSNR $\uparrow$    & SSIM (\%) $\uparrow$   & MSE $\downarrow$      \\ \midrule
			Pix2Pix \cite{isola2017image}          & 26.7000 & 88.56 & 0.0037 & 6.0860 & 0.0480 & 0.0227 & 47.0000 & 10.2000   & 91.08   & 0.0027   \\
			SceneTextEraser \cite{nakamura2017scene} & 25.4700 & 90.14 & 0.0047 & 6.0069 & 0.0533 & 0.0296 & 10.2000 & 25.4000   & 90.12   & 0.0065   \\ 
   EnsNet  \cite{zhang2019ensnet}         & 29.5400 & 92.74     & 0.0024 & 4.1600 & 0.0307  & 0.0136 & 44.4000  & 37.3600 & 96.44     & 0.0021 \\  
   EraseNet  \cite{liu2020erasenet}        & 32.3000 & 95.42     & 0.0015 & 3.0174 & 0.0160  & 0.0090 & 8.5000   & 38.3200 & 97.67     & \underline{0.0002} \\  
   Tang et al.  \cite{tang2021stroke}  & 35.3400 & 96.24     & 0.0009 &- &- &- &- & 38.6000 & 97.55     & \underline{0.0002} \\
   Jiang et al.  \cite{jiang2022self} &34.1400 & 89.15 &- &- &- &- &- &- & - &- \\
   CTRNet  \cite{liu2022don} &\underline{35.8500} &\underline{97.40}  &0.0009 &- &- &-  &\underline{3.3000} &41.2800 &98.50 &{0.0002} \\
   PSSTRNet  \cite{lyu2022psstrnet} &34.6500 &96.75 &0.0014 &\underline{1.7161} &0.0135 &0.0074 &9.3000 &39.2500 &98.15 &\underline{0.0002} \\
   MBE  \cite{hou2022multi} &35.0300 &97.31 &- &2.0594 &0.01282 &0.0088 &- &\textbf{43.8500} &\textbf{98.64} &- \\
   SAEN  \cite{du2023modeling}  &34.7500 &96.53 &\underline{0.0007} &1.9800 &0.0125 &0.0073 &-  &38.6300 &98.27 &0.0003 \\
   PERT  \cite{wang2023real} &33.6200 &97.00 &0.0013 &2.1850 &0.0135 &0.0088 & 7.6000 &39.4000 &97.87 &\underline{0.0002} \\
PEN  \cite{du2023progressive} &35.7200 &96.68 &\textbf{0.0005} &1.9500 &\underline{0.0071} &\textbf{0.0020} &3.9000 &38.8700 &97.83 &0.0003 \\
FetNet  \cite{lyu2023fetnet}        & 34.6500 & 96.75     & 0.0014 & \underline{1.7161} & 0.0135  & 0.0074 & 10.5000  & 39.1400 & 97.97     & \underline{0.0002}  \\ 
ViTEraser  \cite{peng2023viteraser}     & \textbf{37.1100} & \textbf{97.61}     & \textbf{0.0005} & \textbf{1.7000} & \textbf{0.0066} & \underline{0.0035} & \textbf{0.7680} & \underline{42.9700} & \underline{98.55}     & \textbf{0.000092} \\ \bottomrule
		\end{tabular}
        \vspace{-15pt}
	\end{table*}

\begin{figure}[t]
 \setlength{\abovecaptionskip}{0cm} %调整图片标题与图距离
\begin{center}
\includegraphics[width=0.4\textwidth]{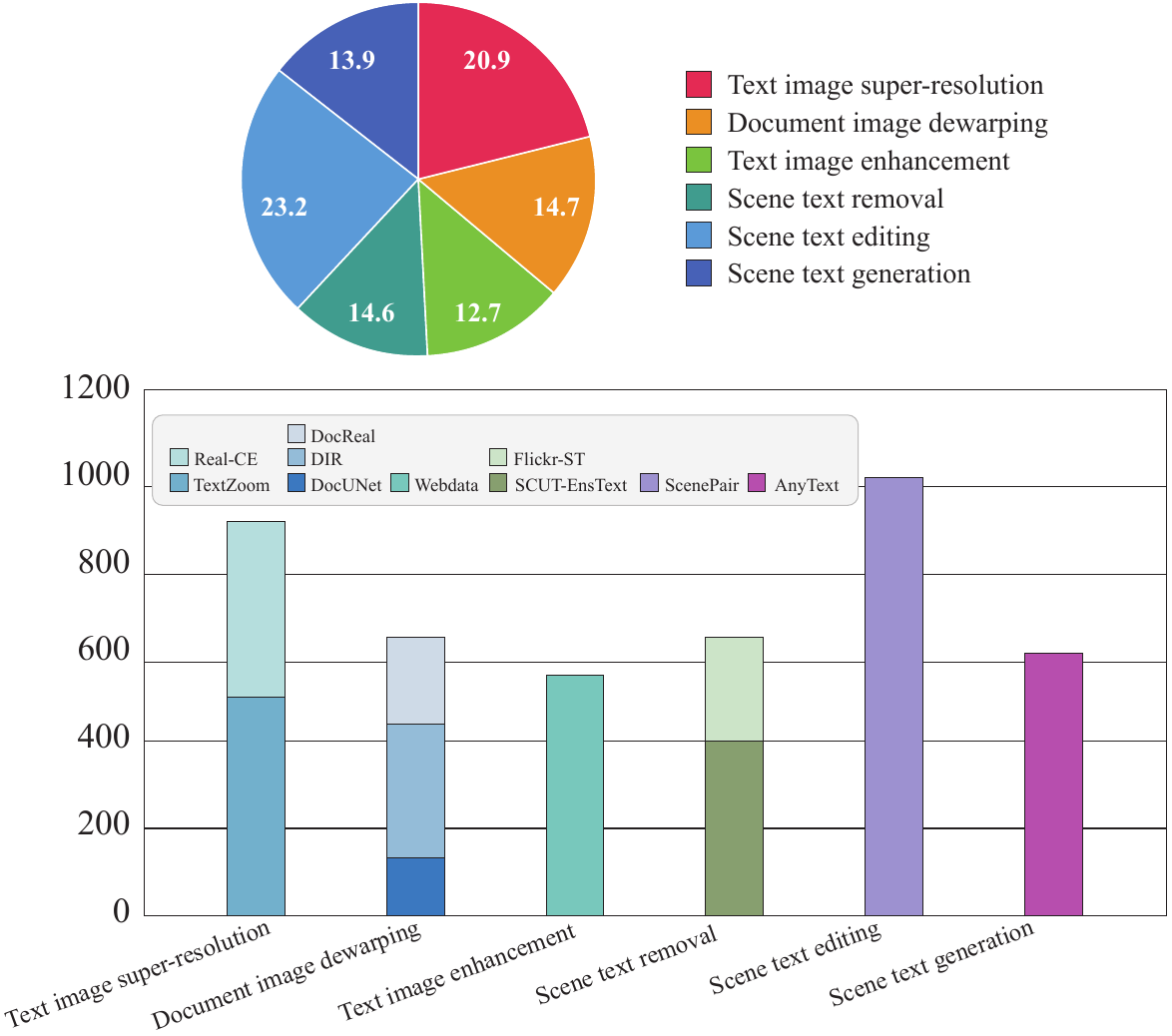}
\hfill
\end{center}
\caption{Statistical overview of our VTPBench. \textbf{Top}: Task type of VTPBench. \textbf{Bottom}: The number of samples and distribution of each task.} 
% \vspace{-15pt}
\label{fig:data_sta}
\end{figure}

\textbf{Scene Text Generation.} For scene text generation (STG) evaluation, web-sourced data comprising images and corresponding caption prompts are collected and filtered. For example, MARIO-Eval \cite{chen2023textdiffuser} gathers data from LAION \cite{schuhmann2021laion} and employs OCR tools to extract text-rich samples. AnyText \cite{tuo2023anytext} collects multilingual data, including Chinese, Korean, and other languages. Similar to scene text editing (STE), OCR-related metrics are used to assess text rendering readability, while Fréchet Inception Distance (FID) evaluates the similarity between synthetic and real-world images.

\subsection{VTPBench and VTPScore}
We propose VTPBench, a multi-task benchmark comprising 4,305 samples across six sub-tasks, specifically designed for evaluating visual text processing.

\textbf{Data Construction.} As discussed in Section \ref{sec:data}, we carefully choose some representative data to construct our VTPBench. An overview of its statistical information is provided in Figure \ref{fig:data_sta}. During our selection process, we filter out extremely broken or severely damaged samples. The data can be totally accessed.\footnote{\url{https://github.com/shuyansy/Survey-of-Visual-Text-Processing}}

\textbf{Settings.} Within the MLLM family, we select GPT-4o \cite{gpt4o} as the base model for unified visual text processing evaluation, as it has demonstrated exceptional visual-language understanding and strong low-level vision perception capabilities. Since reference-based evaluation is generally more reliable than reference-free evaluation, we simultaneously provide GPT-4o with both the predicted result and the corresponding ground truth label from the dataset. We evaluate more than 20 open-source baselines across various visual text processing tasks using their official model weights on VTPBench.

\begin{figure*}[t]
 \setlength{\abovecaptionskip}{0cm} %调整图片标题与图距离
\begin{center}
\includegraphics[width=1\textwidth]{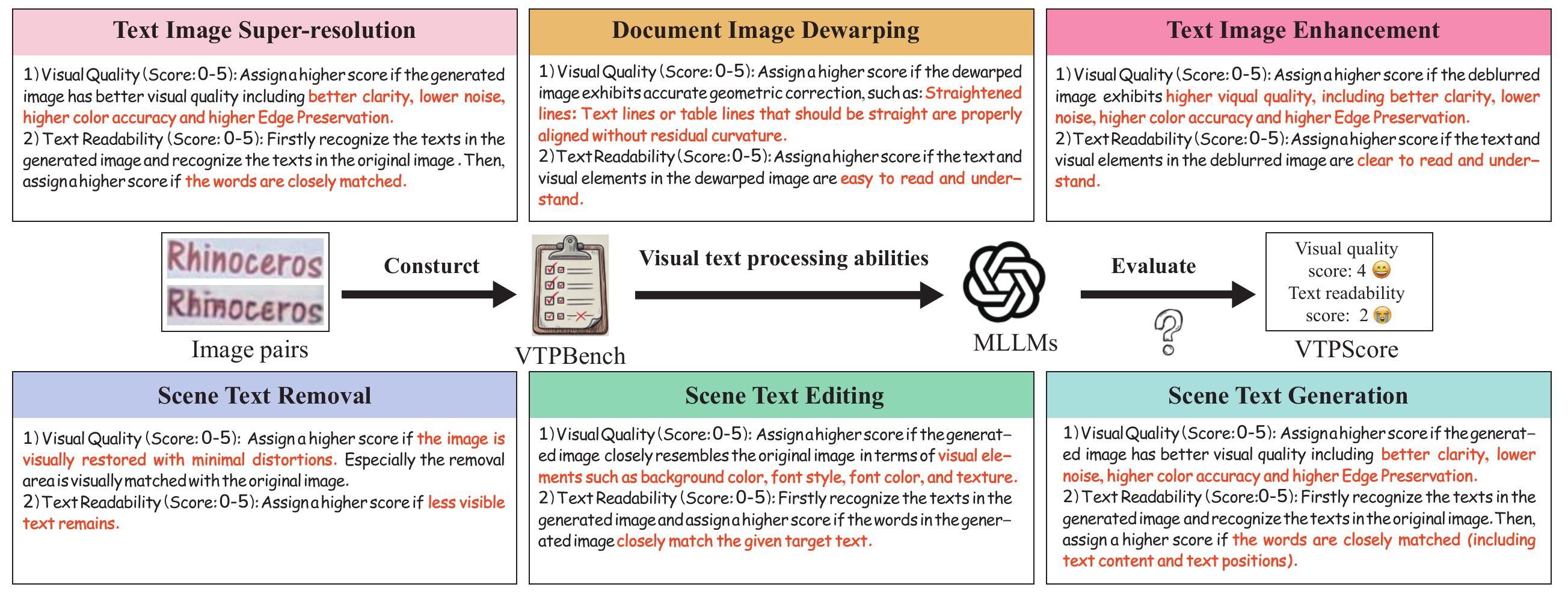}
\hfill
\end{center}
\vspace{-10pt}
\caption{The overview of VTPBench evaluation and details of prompt design (visual quality and text readability) for visual text processing tasks.}
\vspace{-10pt}
\label{fig:prompt}
\end{figure*}

\textbf{VTPScore Evaluation.} Due to the significant gap between different visual text processing tasks, achieving a unified evaluation remains a challenge. To address this, we propose VTPScore, which standardizes evaluation across two key perspectives: \textbf{visual quality} and \textbf{visual text readability}. Additionally, we carefully design task-specific prompts to ensure accurate evaluation for each subtask, as demonstrated in Figure \ref{fig:prompt}.

Visual quality encompasses various aspects of visual elements. In our evaluation, we primarily assess image clarity and blurriness for super-resolution and enhancement tasks. Additionally, we emphasize style consistency between the source and target images. For the dewarping task, we ensure that the model prioritizes geometric accuracy, including margin alignment, shape preservation, and straight-line integrity. In scene text removal and text generation, rather than focusing on overall image quality, we evaluate the quality of the manipulated text region. For instance, the processed area should exhibit minimal artifacts, particularly along the edges. To evaluate visual text readability, we require the MLLM to first recognize text from both the predicted image and the ground truth labels, then compare their matching accuracy. Due to the strong and flexible OCR capabilities of GPT-4o, reliable visual text readability can be assessed without the need for additional OCR modules.

Moreover, another  challenge is extracting scores from MLLM responses, as they often generate explanatory language alongside numerical evaluations. To address this, we design a structured answer prompt that instructs the model to output scores in JSON format. Specifically, VTPScore is computed as the sum of the visual quality score and the visual text readability score, both ranging from 0 to 5.

\begin{table}[t]
\renewcommand{\arraystretch}{1}
\centering
\caption{Scene text editing performance on ScenePair and Tamper-Scene.}
\label{tab:tamper}
\vspace{-10pt}
\resizebox{1\linewidth}{!}{
\begin{tabular}{cccccc}
\toprule
\multirow{2}{*}{Methods} &
\multicolumn{4}{c}{ScenePair}
&
\multicolumn{1}{c}{Tamper-Scene} \\
\cmidrule(r){2-5} \cmidrule(r){6-6}
                         & MSE $\downarrow$    & PSNR $\uparrow$  & SSIM $\uparrow$   & FID $\downarrow$   & Acc $\uparrow$       \\ \midrule
SRNet  \cite{wu2019editing}                    & 5.61 & 14.08 & 26.66 & 49.22  & 39.96      \\
MOSTEL  \cite{qu2023exploring}                   & \underline{5.19} & \underline{14.46} & 27.45 & \underline{49.19} & \textbf{76.79}       \\
DiffSTE  \cite{ji2023improving}                   &6.11 & 13.44 & 26.85& 120.34 & -      \\
TextDiffuser  \cite{chen2023textdiffuser}                   &5.75 & 13.96 & 27.02& 57.01 & -      \\
AnyText  \cite{tuo2023anytext}                   &6.19 & 13.66 & \underline{30.73}& 51.79 & -     \\
TextCtrl  \cite{zeng2024textctrl}                   &\textbf{4.47} & \textbf{14.99} & \textbf{37.56}& \textbf{43.78} & \underline{74.17}       \\
\bottomrule
\end{tabular}
}
\vspace{-10pt}
\end{table}

\begin{table}[t]
\renewcommand{\arraystretch}{1}
\centering
\caption{Scene text generation performance on AnyText.}
\label{tab:anytext}
\vspace{-10pt}
\resizebox{1\linewidth}{!}{
\begin{tabular}{ccccccc}
\toprule
\multirow{2}{*}{Methods} &
\multicolumn{3}{c}{English} &
\multicolumn{3}{c}{Chinese} \\
\cmidrule(r){2-4} \cmidrule(r){5-7}
                         & Sen. ACC $\uparrow$  & NED $\uparrow$  & FID $\downarrow$  & Sen. ACC $\uparrow$  & NED $\uparrow$  & FID $\downarrow$  \\ \midrule
ControlNet  \cite{controlnet}    & 0.5837 & 0.8015 & 45.41  & 0.3620 & 0.6227 & 41.86 \\
TextDiffuser  \cite{chen2023textdiffuser} & 0.5921 & 0.7951 & 41.31 & 0.0605 & 0.1262 & 53.37 \\
GlyphControl  \cite{yang2023glyphcontrol}  & 0.5262 &0.7529  & 43.10 & 0.0454 & 0.1017 & 49.51 \\
AnyText  \cite{tuo2023anytext}  & \underline{0.7239} & \underline{0.8760} & \underline{33.54} & \underline{0.6923} & \underline{0.8396} & \underline{31.58} \\
AnyText2 \cite{tuo2024anytext2} &\textbf{0.8175} &\textbf{0.9193} &\textbf{27.87} &\textbf{0.7250} &\textbf{0.8529} &\textbf{24.32} \\
\bottomrule
\end{tabular}
}
% \vspace{-15pt}
\end{table}

\subsection{Main Results}

In this section, we present a comprehensive performance analysis of the approaches evaluated on existing benchmarks and our proposed VTPBench (See Table \ref{tab:sr}). Additionally, we assess the effectiveness of VTPScore by comparing it with human evaluation results.

\textbf{Performance Comparison.} The effectiveness of text image super-resolution methods is presented in Table \ref{tab:TextZoom}. Notably, LEMMA emerges as the top-performing method in both text image super-resolution and VTPBench, achieving the highest scores in visual quality evaluation (3.07) and visual text readability (4.16). LEMMA incorporates fine-grained semantic character information, enhancing its ability to model character restoration effectively.

For performance benchmarking in document image dewarping, a quantitative comparison of various DID methods is presented in Table \ref{tab:docunet}. Early methods, such as DewarpNet, focus on 3D reconstruction and flattening, making it challenging to capture fine details. More recent approaches emphasize both global features (e.g., foreground images) and local features (e.g., text lines, control points, and layout), leading to significant advancements. Notably, Li et al. \cite{Li2023LayoutAwareSD} achieve leading results on DocUnet and DIR300, while UVDoc achieves the best results on VTPBench, as it enables the simultaneous prediction of both 3D shape and 2D flow mapping.

\begin{table}[t]
\centering
\caption{ Experimental results of VTPBench towards different visual text processing tasks, including text image super-resolution (TISR), document image dewarping (DID), text image enhancement (TIE), scene text removal (STR), scene text editing (STE) and scene text generation (STG). ``VS" and ``TS" denote visual quality score and visual text readability score, respectively. Bold denotes the \textbf{best} result.}
\vspace{-10pt}
\renewcommand{\arraystretch}{1.15}
\resizebox{0.47\textwidth}{!}{%
    \begin{tabular}{cccccc}
        \toprule
        \textbf{Task} & \textbf{Methods}  &\textbf{VS} & \textbf{TS} & \textbf{VTPScore} &  \textbf{HumanScore} \\
        \midrule      
        \multirow{4}{*}{TISR} 
        & TSRN \cite{Wang2020Scene} & 2.87 & 3.82 &6.70 &3.58  \\
        & TBSRN \cite{Chen2021SceneTT} & 2.97 & 3.98 &6.96 &7.25   \\
        & Text-Gestalt \cite{Chen2021TextGS} & 2.94 & 4.01 &6.96 &7.40 \\
        & LEMMA \cite{guo2023towards} & \textbf{3.07} & \textbf{4.16} &\textbf{7.24} &\textbf{8.20} \\
        \hline
        \multirow{4}{*}{DID} 
        & DDCP\cite{Xie2022DocumentDW} & 2.78 & 2.08 &4.86 &2.40  \\
        & DewarpNet \cite{Das2019DewarpNetSD} & 2.98 & 2.68 &5.66 &6.54   \\
        & DocGeoNet \cite{Feng2022GeometricRL} & 3.05 & 2.70 &5.75 &6.68 \\
        & UVDOC \cite{verhoeven2023uvdoc} & \textbf{3.20} & \textbf{2.88} &\textbf{6.08} &\textbf{7.90} \\
        \hline
        \multirow{3}{*}{TIE} 
        & DocDiff \cite{Yang2023DocDiffDE} & 1.90 & 1.89 &3.79 &4.48   \\
        & NAF-DPM\cite{cicchetti2024nafdpm} & 2.55 & 2.62 &5.17&5.02 \\
        & DocRes \cite{zhang2024docres} & \textbf{2.75} & \textbf{2.74} &\textbf{5.49} &\textbf{6.20} \\
        \hline
        \multirow{4}{*}{STR} 
        & EraseNet \cite{liu2020erasenet} & 3.91 & 3.87 &7.78 &6.20   \\
        & CTRNet\cite{liu2022don} & 3.03 & 3.52 &6.55&6.06 \\
        & Pert\cite{wang2021pert} & 3.97 & 3.89 &7.86 &7.58 \\
        & ViTEraser \cite{peng2023viteraser} & \textbf{4.48} & \textbf{4.35} &\textbf{8.83} &\textbf{8.90} \\
        \hline
        \multirow{5}{*}{STE} 
        & MOSTEL \cite{tang2021stroke} & 4.20 & 2.61 &6.81 &6.35   \\
        & DiffSTE\cite{ji2023improving} & 3.68 & 2.79 &6.47&5.46 \\
        & TextDiffuser\cite{chen2023textdiffuser} & 3.77 & 3.45 &7.22 &7.05 \\
        & AnyText\cite{tuo2023anytext} & 3.78 & 3.45 &7.23 &7.20 \\
        & TextCtrl \cite{zeng2024textctrl} & \textbf{4.48} & \textbf{4.54} &\textbf{9.02} &\textbf{9.50} \\
        \hline
        \multirow{3}{*}{STG} 
        & GlyphControl\cite{yang2023glyphcontrol} & 3.22 & 2.10 &5.32&5.30 \\
        & TextDiffuser\cite{chen2023textdiffuser} & 3.60 & 2.17 &5.77 &6.25 \\
        & AnyText\cite{tuo2023anytext} & \textbf{3.70} & \textbf{2.93} &\textbf{6.64} &\textbf{7.36} \\
  
        \bottomrule
    \end{tabular}%
}
\vspace{-10pt}
\label{tab:sr}
\end{table}

% \begin{table}[h]
% \center
%  \caption{Experimental results of text image denoising methods.}
%  \vspace{-10pt}
% \renewcommand{\arraystretch}{1.15}
%     \begin{tabular}{>{\kern-0.5\tabcolsep}c|cccc<{\kern-0.5\tabcolsep}}
%         \toprule
%         \textbf{Methods}  &\textbf{VS} & \textbf{TS} & \textbf{VTPScore} &  \textbf{HumanScore} \\
%         \midrule      
%         DocDifff \cite{Yang2023DocDiffDE} & 1.90 & 1.89 &3.79 &4.48   \\
%         NAF-DPM\cite{cicchetti2024nafdpm} & 2.55 & 2.62 &5.17&5.02 \\
%         DocRes \cite{zhang2024docres} & \textbf{2.75} & \textbf{2.74} &\textbf{5.49} &\textbf{6.20} \\
%         \bottomrule
%     \end{tabular}
% \label{tab:tid}
% \end{table}

% \begin{table}[h]
% \center
%  \caption{Experimental results of scene text removal methods.}
%  \vspace{-10pt}
% \renewcommand{\arraystretch}{1.15}
%     \begin{tabular}{>{\kern-0.5\tabcolsep}c|cccc<{\kern-0.5\tabcolsep}}
%         \toprule
%         \textbf{Methods}  &\textbf{VS} & \textbf{TS} & \textbf{VTPScore} &  \textbf{HumanScore} \\
%         \midrule      
%         EraseNet \cite{liu2020erasenet} & 3.91 & 3.87 &7.78 &6.20   \\
%         CTRNet\cite{liu2022don} & 3.03 & 3.52 &6.55&6.06 \\
%         Pert\cite{wang2021pert} & 3.97 & 3.89 &7.86 &7.58 \\
%         ViTEraser \cite{peng2023viteraser} & \textbf{4.48} & \textbf{4.35} &\textbf{8.83} &\textbf{8.90} \\
%         \bottomrule
%     \end{tabular}
% \label{tab:str}
% \end{table}

Compared to text image super-resolution and dewarping, text image enhancement remains a more challenging task. The strongest model, DocRes, achieves only 5.49 in VTPBench. We attribute this to the significant gap between the test data and the training data, which affects the performance of the model.

The effectiveness of various scene text removal methods is showcased on Table \ref{tab:enstext}.  On SCUT-Syn dataset, MBE outperforms other methods in PSNR and SSIM, benefiting from its ensemble strategy. In contrast, ViTEraser achieves the best performance on SCUT-EnsText and VTPBench across most metrics. This superior performance can be attributed to its self-training scheme for pre-training, which enables the model to learn from real-world data more effectively.

% \begin{table}[h]
% \center
%  \caption{Experimental results of scene text editing methods.}
%  \vspace{-10pt}
% \renewcommand{\arraystretch}{1.15}
%     \begin{tabular}{>{\kern-0.5\tabcolsep}c|cccc<{\kern-0.5\tabcolsep}}
%         \toprule
%         \textbf{Methods}  &\textbf{VS} & \textbf{TS} & \textbf{VTPScore} &  \textbf{HumanScore} \\
%         \midrule      
%         MOSTEL \cite{tang2021stroke} & 4.20 & 2.61 &6.81 &6.35   \\
%         DiffSTE\cite{ji2023improving} & 3.68 & 2.79 &6.47&5.46 \\
%         TextDiffuser\cite{chen2023textdiffuser} & 3.77 & 3.45 &7.22 &7.05 \\
%         AnyText\cite{tuo2023anytext} & 3.78 & 3.45 &7.23 &7.20 \\
%         TextCtrl \cite{zeng2024textctrl} & \textbf{4.48} & \textbf{4.54} &\textbf{9.02} &\textbf{9.50} \\
%         \bottomrule
%     \end{tabular}
% \label{tab:ste}
% \end{table}

% \begin{table}[h]
% \center
%  \caption{Experimental results of scene text generation methods.}
%  \vspace{-10pt}
% \renewcommand{\arraystretch}{1.15}
%     \begin{tabular}{>{\kern-0.5\tabcolsep}c|cccc<{\kern-0.5\tabcolsep}}
%         \toprule
%         \textbf{Methods}  &\textbf{VS} & \textbf{TS} & \textbf{VTPScore} &  \textbf{HumanScore} \\
%         \midrule      
%         GlyphControl\cite{yang2023glyphcontrol} & 3.22 & 2.10 &5.32&5.30 \\
%         TextDiffuser\cite{chen2023textdiffuser} & 3.60 & 2.17 &5.77 &6.25 \\
%         AnyText\cite{tuo2023anytext} & \textbf{3.70} & \textbf{2.93} &\textbf{6.64} &\textbf{7.36} \\
%         \bottomrule
%     \end{tabular}
% \label{tab:stg}
% \end{table}

It is important to recognize that many scene text editing approaches primarily rely on synthetic datasets for both training and testing, potentially introducing biases in performance evaluation. As detailed in Table \ref{tab:tamper}, early methods such as MOSTEL focus on general style transfer but struggle with fine-grained text rendering. More recent approaches leverage advanced diffusion models, significantly improving visual similarity and reducing spelling errors. Among these methods, TextCtrl demonstrates superior performance.

The results of the scene text generation task are shown in Table \ref{tab:anytext}. It can be seen that AnyText achieves state-of-the-art performance, benefiting from its strong generalization capabilities in multilingual and multi-oriented visual text generation. Notably, on VTPBench evaluation, both TextDiffuser and AnyText demonstrate promising results in scene text editing and generation, highlighting the potential for developing a unified multi-task model.

\textbf{Human Evaluation.} To assess the gap between VTPScore and human performance in visual text processing, we conducted a human study where ten participants assessed VTPBench using the same criteria as VTPScore. Specifically, participants rate visual quality and text readability on a 0–5 scale. The participants come from diverse fields and possess extensive experience in relevant domains. Before the test, participants undergo a brief training session to familiarize themselves with the objectives of different tasks and establish a unified evaluation protocol through discussion. Afterward, they complete the test, and we record their average scores as the final human performance benchmark. To mitigate fatigue, each participant labels up to 50 images per day, ensuring that each sample is evaluated by at least three participants. The results, presented in Table \ref{tab:sr}, demonstrate a high consistency between VTPScore and human evaluation.

\section{Open Challenges}
\label{section5}
Despite recent advancements in visual text processing, numerous challenges remain unresolved. This section highlights key open issues and potential future directions for further research.

\subsection{Training Data}
The scarcity of labeled real-world training data remains a significant bottleneck in the development of visual text image processing methods. For instance, acquiring paired source and target data with consistent source styles presents a notable challenge in scene text editing tasks. Additionally, training data obtained from the web are frequently contaminated with noise and subject to scene constraints. For example, benchmarks such as LAION \cite{schuhmann2021laion}, which are utilized for text image generation, predominantly comprise poster and web data, lacking sufficient natural scene images.   Consequently, the compilation of comprehensive and high-quality datasets remains an unresolved issue in the field.

Future trends may pivot on optimizing the trade-off between dataset quality and quantity. A key question is whether models perform better with weaker supervision across extensive datasets or with stronger supervision derived from smaller and high-quality datasets. If weaker supervision proves beneficial, advancements in self-supervised and semi-supervised learning could enable models to leverage vast amounts of unannotated data more effectively. On the other hand, if strong supervision is preferred, improving model generalizability in data-scarce scenarios will be crucial. This could be achieved through auxiliary techniques such as domain adaptation, allowing models to transfer knowledge across different datasets and real-world conditions.

% \subsection{Evaluation Metrics}
% As discussed earlier, prevalent visual text processing techniques often rely on Image-Eval and Det/Rec-Eval metrics for assessment. However, the applicability of certain Image-Eval metrics, such as PSNR and SSIM, is limited due to the absence of ground truth pairs. Furthermore, general image and video quality metrics like FID may not be entirely appropriate for text image evaluations due to the domain shift from natural images to visual text images. On the other hand, Det/Rec-Eval metrics can lead to skewed comparisons as various detectors or recognizers are employed across different methodologies. Additionally, the selection of hyperparameters and data augmentation techniques can significantly influence the outcomes.

A promising direction for advancement in this field is the development of enhanced metrics tailored for the text image domain. These metrics ought to be versatile, accommodating a wide range of multilingual text types (such as English and Chinese), various text shapes (including horizontal and oriented texts), and diverse environments (like posters and street scenes). Additionally, these metrics should closely correlate with human judgment, facilitating accelerated and autonomous progress in methodological development with minimal human intervention.

\subsection{Efficiency and Complexity}
 Efficiency remains a critical issue for visual text processing techniques. While many studies highlight substantial accuracy improvements, they often overlook reporting on model complexity (FLOPS) and inference speed (FPS). As a result, most existing methods struggle to achieve an optimal balance between accuracy and computational efficiency. This is largely due to the inherent architectural complexities, such as the self-attention mechanism in Transformers leading to intricate calculations or the slow sampling rates in diffusion models \cite{ho2020denoising} that impede swift inference. Additionally, certain multi-stage approaches fail to consider overall system efficiency, limiting their practical applicability. For example, text removal methods should seamlessly incorporate a text detection mechanism to generate text masks.

A practical approach to enhance efficiency is the development of novel, streamlined architectures that reduce the time required for each denoising step in diffusion models \cite{li2023snapfusion} and decreasing computational complexity in Transformers. Techniques like model distillation also strive to improve efficiency. Furthermore, the use of end-to-end architectures can eliminate the need for auxiliary modules, streamlining the process further.

\subsection{Extension to Videos}
While 2D visual text image processing has advanced significantly because of technological progress and data availability, the evolution in higher-dimensional contexts, such as video, remains relatively limited. The only video text processing method is STRIVE \cite{subramanian2021strive}, which aims for video scene text editing. The challenges in video-based visual text processing are manifold. Firstly, data availability and quality present substantial challenges. Although there is an abundance of raw video data, annotating this data to capture motion and temporal dependencies is a complex task. The lack of high-quality annotated data restricts the development of robust and generalizable models for processing visual text in videos. Secondly, the complexity of network architecture design poses another hurdle. Higher-dimensional data cannot be handled as simply as 2D images, which rely on discrete pixel values. Instead, they demand more sophisticated representations to manage long-range information crucial for interpreting temporal dynamics in videos and spatial relationships.

Future research should prioritize leveraging the vast amount of online videos to build high-quality video datasets. This endeavor will require substantial engineering efforts and the development of dedicated automatic curation tools to enable efficient annotation and scalable dataset construction. In addition, it is crucial to design video text processing architectures that can effectively handle the high-dimensional nature of video data—similar to general video understanding models\cite{shu2024video}—while also addressing the diverse characteristics of textual information.

\subsection{Unified Framework}
Contemporary research in visual text processing often focuses on frameworks designed for isolated tasks, overlooking their interconnected nature. However, in real-world applications, users typically have multifaceted needs. For instance, within a single scene text image, a user may require simultaneous text removal, editing, and generation. Moreover, user interests often extend beyond textual elements to include various objects within the scene. A model capable of processing text but lacking an understanding of the broader scene composition remains significantly limited.

Future research should focus on breaking down the barriers between interrelated visual text processing tasks, aiming to develop a cohesive and adaptable framework capable of handling multiple tasks within a unified system. For instance, DocRes \cite{zhang2024docres} has been proposed as a generalist model that unifies five document image restoration tasks: dewarping, de-shadowing, appearance enhancement, deblurring, and binarization. In the text image manipulation field, UPOCR \cite{peng2023upocr} is a unified framework to address text removal, segmentation, and tamper detection. However, additional tasks should be incorporated to develop a model capable of simultaneously enhancing, modifying, and synthesizing both text and common objects within images. To achieve this goal, it is crucial to leverage the perception and understanding capabilities of multimodal large language models (MLLMs) to enhance and manipulate visual text effectively.

\subsection{MLLMs-based System} Beyond their strong performance in visual language understanding, MLLMs have shown remarkable capabilities in low-level visual perception and manipulation. Consequently, a natural approach is to harness the power of MLLMs for visual text processing tasks.

This work represents the first attempt to utilize MLLMs for unified visual text processing evaluation. However, further exploration is needed to develop a visual text-specific foundation model framework. First, adapting language centric MLLMs to vision centric perception requires careful design. The mainstream approach involves constructing a de-tokenizer to transform image tokens into real image tokens. However, a key concern is the quality of restored images, particularly for text-rich data, which contains fine-grained semantic and textural information. Another challenge lies in scaling up visual-language-instructed data for training, where synthetic techniques can be incorporated to enhance dataset diversity.

% \subsection{User-friendly Interaction}
% Current visual text processing approaches typically address all text regions in an image. However, users often need to tailor modifications to their individual requirements. To date, few studies in the fields of text removal \cite{mitani2023selective} and editing \cite{chen2023diffute} employ conditional models or extensive language models to facilitate precise content and style transfer. Despite this, such research is in its initial phase. The method of integrating diverse prompts or inputs for customized processing in various tasks presents a significant research opportunity.

% The emergence of integrated visual models \cite{kirillov2023segment} and multimodal language-vision frameworks \cite{mao2023gpteval} has enabled the processing of diverse textual and visual prompts. Models such as SAM \cite{kirillov2023segment} now support visual prompts like points or bounding boxes to identify areas of interest, and advanced language models can interpret user-provided natural language instructions to derive precise image processing commands. Additionally, methods like in-context learning \cite{dong2022survey} and instruction tuning \cite{liu2023visual} are instrumental in translating personalized user instructions into specific visual text image processing results.

\section{Conclusion}
\label{section7}
In this paper, we provide a comprehensive review of recent advancements in visual text processing tasks, presenting the first specialized survey in this domain to the best of our knowledge. Specifically, we examine the types of text features used in seminal works and discuss various learning paradigms that drive progress in the field. Additionally, we introduce VTPBench and VTPScore, which provide a unified evaluation framework for multiple visual text processing tasks. Finally, we share our perspectives on open challenges and future directions of visual text image processing. We hope this work provides valuable insights for the research community in methodological advancements, benchmark evaluation, and future developments.

% hierarchical taxonomy that encompasses areas ranging from image enhancement and restoration to image manipulation, followed by specific learning paradigms. Additionally, we delve into text features closely related to the mainstream methods, including structure, stroke, semantics, style, and spatial context. Furthermore, we summarize the datasets for benchmarking and tabulate and compare the performance of existing approaches in various visual text processing tasks. 

% or


% \appendices

% % you can choose not to have a title for an appendix
% % if you want by leaving the argument blank
% \section{}

% % use section* for acknowledgment
% \ifCLASSOPTIONcompsoc
%   % The Computer Society usually uses the plural form
%   \section*{Acknowledgments}
% \else
%   % regular IEEE prefers the singular form
% \fi

% This work is supported by the National Natural Science Foundation of China (Grant NO 62376266 and 62406318).

\ifCLASSOPTIONcaptionsoff
  \newpage
\fi



%\bibliographystyle{IEEEtran}


% \end{thebibliography}

\bibliographystyle{IEEEtran}  
\bibliography{reference}  

\end{document}